\documentclass[letterpaper]{article} 
\usepackage{aaai24}  
\usepackage{times}  
\usepackage{helvet}  
\usepackage{courier}  
\usepackage[hyphens]{url}  
\usepackage{graphicx} 
\usepackage{natbib}  
\usepackage{caption} 
\usepackage{algorithm}
\usepackage{algorithmic}

\usepackage{newfloat}
\usepackage{listings}
\DeclareCaptionStyle{ruled}{labelfont=normalfont,labelsep=colon,strut=off} 
\floatstyle{ruled}
\newfloat{listing}{tb}{lst}{}
\floatname{listing}{Listing}
\title{Contextual Pre-planning on Reward Machine Abstractions for Enhanced Transfer in Deep Reinforcement Learning}
\author {
    Guy Azran\textsuperscript{\rm 1},
    Mohamad H. Danesh\textsuperscript{\rm 2},
    Stefano V. Albrecht\textsuperscript{\rm 3},
    Sarah Keren\textsuperscript{\rm 1}
}
\affiliations{
    \textsuperscript{\rm 1}Taub Faculty of Computer Science, Technion -- Israel Institute of Technology\\
    \textsuperscript{\rm 2}School of Computer Science, McGill University\\
    \textsuperscript{\rm 3}School of Informatics, University of Edinburgh\\
    guy.azran@campus.technion.ac.il, mo.danesh@mail.mcgill.ca, 
    s.albrecht@ed.ac.uk, 
    sarahk@cs.technion.ac.il
    
}

\newcount\Comments
\usepackage{xcolor}
\definecolor{darkgreen}{rgb}{0,0.7,0}

\newcommand{\kibitz}[2]{\ifnum\Comments=1{\color{#1}{#2}}\fi}

\usepackage{xspace}

\newcommand{\tuple}[1]{\ensuremath{\left\langle #1 \right\rangle}\xspace}
\newcommand{\real}[0]{\ensuremath{\mathbb{R}}\xspace}
\newcommand{\expectation}[0]{\ensuremath{\mathbb{E}}\xspace}

\newcommand{\history}[0]{\ensuremath{h}\xspace}

\newcommand{\TL}[0]{TL\xspace}
\newcommand{\DRL}[0]{DRL\xspace}

\newcommand{\stateSpace}[0]{\ensuremath{S}\xspace}
\newcommand{\actionSpace}[0]{\ensuremath{A}\xspace}
\newcommand{\transitionFunc}[0]{\ensuremath{T}\xspace}
\newcommand{\rewardFunc}[0]{\ensuremath{R}\xspace}
\newcommand{\discountFactor}[0]{\ensuremath{\gamma}\xspace}

\newcommand{\mdpTup}[0]{\tuple{\stateSpace, \actionSpace, \transitionFunc, \rewardFunc, \discountFactor}\xspace}

\newcommand{\mdpSym}[0]{\ensuremath{M}\xspace}
\newcommand{\stateSym}[0]{\ensuremath{s}\xspace}
\newcommand{\actionSym}[0]{\ensuremath{a}\xspace}
\newcommand{\rewardSym}[0]{\ensuremath{r}\xspace}
\newcommand{\timeSym}[0]{\ensuremath{t}\xspace}

\newcommand{\MDP}[0]{MDP\xspace}

\newcommand{\policyFunc}[0]{\ensuremath{\pi}\xspace}
\newcommand{\optimalPolicy}[0]{\ensuremath{\policyFunc^*}\xspace}

\newcommand{\returnFunc}[0]{\ensuremath{J}\xspace}

\newcommand{\valueFunc}[0]{\ensuremath{V}\xspace}
\newcommand{\viIndex}[0]{\ensuremath{k}\xspace}

\newcommand{\SAS}[0]{\ensuremath{\stateSpace\times\actionSpace\times\stateSpace}\xspace}

\newcommand{\qFunc}[0]{\ensuremath{Q}\xspace}

\newcommand{\qParams}[0]{\ensuremath{\theta}\xspace}
\newcommand{\qFuncParameterized}[0]{\ensuremath{\qFunc_{\qParams}}\xspace}

\newcommand{\contextSpace}[0]{\ensuremath{C}\xspace}
\newcommand{\mdpFunc}[0]{\ensuremath{\mathcal{\mdpSym}}\xspace}

\newcommand{\contextSym}[0]{\ensuremath{c}\xspace}
\newcommand{\mdpInContextSym}[0]{\ensuremath{\mdpFunc_\contextSym}\xspace}
\newcommand{\contextDistSym}[0]{\ensuremath{\Psi}\xspace}

\newcommand{\CMDP}[0]{CMDP\xspace}

\newcommand{\cmdpTup}[0]{\tuple{\contextSpace, \stateSpace, \actionSpace, \mdpFunc}}
\newcommand{\mdpInContextTup}[0]{\tuple{\stateSpace, \actionSpace, \transitionFunc_\contextSym, \rewardFunc_\contextSym, \discountFactor}\xspace}

\newcommand{\utilityFunc}[0]{\ensuremath{\mathcal{U}}\xspace}
\newcommand{\utilityFuncLabeled}[1]{\ensuremath{\utilityFunc_{#1}}\xspace}

\newcommand{\srcContexts}[0]{\ensuremath{\contextSpace_{\text{src}}}\xspace}
\newcommand{\tgtContexts}[0]{\ensuremath{\contextSpace_{\text{tgt}}}\xspace}
\newcommand{\nSrc}[0]{\ensuremath{N_{\text{src}}}\xspace}
\newcommand{\nTgt}[0]{\ensuremath{N_{\text{tgt}}}\xspace}

\newcommand{\js}[0]{JS\xspace}
\newcommand{\jsUtil}[0]{\utilityFuncLabeled{\js}\xspace}

\newcommand{\ttt}[0]{TT\xspace}
\newcommand{\tttUtil}[0]{\utilityFuncLabeled{\ttt}}
\newcommand{\thresh}[0]{\ensuremath{\tau}\xspace}
\newcommand{\auc}[0]{AUC\xspace}
\newcommand{\tttAucUtil}[0]{\utilityFuncLabeled{\ttt_{\auc}}}
\newcommand{\tr}[0]{TR\xspace}
\newcommand{\trUtil}[0]{\utilityFuncLabeled{\tr}}

\newcommand{\tgtPolicy}[0]{\ensuremath{\policyFunc_{\tgtContexts}}\xspace}
\newcommand{\tsfPolicy}[0]{\ensuremath{\policyFunc_{\srcContexts, \tgtContexts}}\xspace}

\newcommand{\rmTerm}[0]{RM\xspace}
\newcommand{\rmAbsStateSpace}[0]{\ensuremath{U}\xspace}
\newcommand{\rmAbsTransFunc}[0]{\ensuremath{\delta_{u}}}
\newcommand{\rmAbsRewardFunc}[0]{\ensuremath{\delta_{r}}}

\newcommand{\rmTup}[0]{\tuple{\rmAbsStateSpace, \rmAbsTransFunc, \rmAbsRewardFunc}}

\newcommand{\rmInitAbsStateSym}[0]{\ensuremath{\rmAbsStateSym_0}\xspace}
\newcommand{\rmPropSet}[0]{\ensuremath{\mathcal{P}}\xspace}
\newcommand{\rmPropSym}[0]{\ensuremath{l}\xspace}
\newcommand{\rmAbsStateSym}[0]{\ensuremath{u}\xspace}

\newcommand{\rmSym}[0]{\ensuremath{\mathfrak{R}}\xspace}

\newcommand{\rmInContextTup}[0]{\tuple{\rmAbsStateSpace^{\contextSym}, \rmAbsTransFunc^{\contextSym}, \rmAbsRewardFunc^{\contextSym}}}

\newcommand{\stateLabelFunc}[0]{\ensuremath{L}\xspace}
\newcommand{\potentialFunc}[0]{\ensuremath{\phi}\xspace}

\newcommand{\cPrep}[0]{C-PREP\xspace}
\newcommand{\cPrepTerm}[0]{Contextual PRE-Planning\xspace}

\newcommand{\rmGenFunc}[0]{\ensuremath{G}\xspace}

\newcommand{\augStateSym}[0]{\ensuremath{\hat{\stateSym}}\xspace}

\newcommand{\itersSym}[0]{\ensuremath{N}\xspace}
\newcommand{\replayBufferSym}[0]{\ensuremath{\mathcal{D}}\xspace}
\newcommand{\qValuesSym}[0]{\ensuremath{q}\xspace}
\newcommand{\qLossFunc}[0]{\ensuremath{l}\xspace}

\newcommand{\algDQN}[0]{DQN\xspace}

\newcommand{\modRS}[0]{RS\xspace}

\newcommand{\modLTL}[0]{LTL\xspace}
\newcommand{\modDTL}[0]{DTL\xspace}

\newcommand{\ctxCTL}[0]{CTL\xspace}
\newcommand{\ctxPCG}[0]{PCG\xspace}

\newcommand{\stateLabelFuncTerm}[0]{transition labeling function\xspace}
\newcommand{\rmGenFuncTerm}[0]{\rmTerm generator function\xspace}
\newcommand{\transitionLabelTerm}[0]{transition label\xspace}

\usepackage{amsmath}
\usepackage{amssymb}
\newtheorem{example}{Example}  
\DeclareMathOperator*{\argmax}{arg\,max}

\usepackage{subcaption}  
\usepackage{cprotect}  

\usepackage{multirow}  
\usepackage{makecell}  
\usepackage{rotating}  
\usepackage{tablefootnote}  
\newcolumntype{?}{!{\vrule width 4\arrayrulewidth}}  

\usepackage[capitalize]{cleveref}  

\usepackage{enumitem}  

\begin{document}

\maketitle

\begin{abstract}

Recent studies show that deep reinforcement learning (\DRL) agents tend to overfit to the task on which they were trained and fail to adapt to minor environment changes.
To expedite learning when transferring to unseen tasks, we propose a novel approach to representing the current task using {\em reward machines} (\rmTerm{}s), state machine abstractions that induce subtasks based on the current task's rewards and dynamics. Our method provides agents with symbolic representations of optimal transitions from their current abstract state and rewards them for achieving these transitions. These representations are shared across tasks, allowing agents to exploit knowledge of previously encountered symbols and transitions, thus enhancing transfer. Empirical results show that our representations improve sample efficiency and few-shot transfer in a variety of domains.

\end{abstract}

\section{Introduction}

Reinforcement learning (RL) methods, especially deep RL (DRL) methods, have shown impressive capabilities in a wide variety of problems \citep{chen_system_2021,schrittwieser_mastering_2020}. However, recent studies show that these algorithms have difficulty adapting to even the slightest variations in the agent's objective or environment dynamics \citep{danesh_out--distribution_2022,agarwal_contrastive_2020, zhang_study_2018,leike_ai_2017}. Adapting quickly to new tasks is imperative in real-world scenarios, such as robotics \citep{dunion_conditional_2023,dunion_temporal_2023} and healthcare \citep{tseng_deep_2017}, where agents reside in a dynamic world with ever-changing objectives and constraints. Consequently, agents require many interactions with the environment to learn to perform new tasks despite having mastered similar ones. The problem is exacerbated for tasks with sparse reward signals \citep{gupta_unpacking_2022} and long-term dependencies between actions \citep{langford_real_2018}.

\begin{example}\label{ex:house-contexts}
    A housekeeper robot learns to do multiple tasks, one of which is to make coffee in a mug. Next, the robot is tasked with making coffee in a glass, something it has never attempted. The two tasks are similar in that they interact with many of the same objects (e.g. coffee, spoon, etc.) and perform identical subtasks (e.g. boil water, fill cup, etc.). The robot is expected to use its experience in making coffee in a mug to learn to achieve the new task more quickly. 
\end{example}

A {\em contextual MDP} (\CMDP) \citep{langford_contextual_2017, hallak_contextual_2015} models settings like \cref{ex:house-contexts} as a collection of tasks in the same environment, where each task is represented by the current {\em context}. \CMDP{}s have been used in recent work that aims to improve {\em zero-shot} transfer capabilities, i.e., solving new tasks after training on a subset of them \citep{benjamins_contextualize_2022, hallak_contextual_2015}. In contrast, we aim to improve {\em few-shot} transfer, in which the agent may continue training on previously unseen tasks with the objective of minimizing the additional training required to achieve desirable performance. 

One of the key challenges when using a \CMDP to model transfer learning settings is finding a concise way to represent the current context while maximizing transfer capabilities. 
For this, we take advantage of {\em reward machines} (\rmTerm{}s) \citep{toro_icarte_using_2018}, state-machine-based abstractions that represent the structure of the reward function and the dynamics of a task and its subtasks. Transitions between abstract states in the \rmTerm occur when certain facts, represented as binary symbols, hold true. As the agent traverses the environment, it keeps track of these facts and its current \rmTerm state. \citet{camacho_reward_2021} used \rmTerm{}s by providing the agent with the current abstract state and showed that this can expedite learning on a single task. In contrast, we leverage \rmTerm{}s to improve transfer to new tasks.

Our novel technique, called {\em \cPrepTerm} (\cPrep), takes as input a \CMDP and an RM generator function 
that represents contextual information through task-specific \rmTerm{} abstractions with shared symbolic representations. Given a task, \cPrep finds an optimal policy in the corresponding \rmTerm abstraction and gives the agent the next desired abstract transition according to that policy as additional input. Furthermore, \cPrep uses the \rmTerm by reshaping the reward function according to abstract state transitions within the \rmTerm, thus highlighting important transitions throughout learning. When transferred to a new task, the agent can exploit abstract transitions that it has encountered during training and needs only to adapt to symbols with which it has not previously interacted.

We empirically evaluate \cPrep in various environments with sparse rewards and varying difficulties. In our experiments, a \algDQN agent \citep{mnih_human-level_2015} is initially trained on a collection of source contexts. Subsequently, we transfer the policy network to a different set of target contexts, where it undergoes further training and evaluation.
We observe an improvement in few-shot as well as zero-shot transfer performance when using \cPrep compared to other context representation methods. The performance gap grows as the problem difficulty increases, with improvements of 22.84\% to 42.31\% in time-to-threshold (few-shot transfer), and from 11.86\% to 36.5\% in jumpstart (zero-shot transfer) for the most complex tasks compared to the next best baseline.

\section{Background}\label{sec:bg}

{\bf Reinforcement learning} (RL) is a method for agent learning through experiencing the world, acting within it, and receiving rewards (both positive and negative) for achieving certain states or state transitions. RL problems commonly model the world as a \textbf{Markov decision process} (MDP) \citep{bellman_markovian_1957} $\mdpSym = \mdpTup$ where \stateSpace is a set of possible states, \actionSpace is a set of agent actions, $\transitionFunc:\SAS \rightarrow [0, 1]$ is the state transition function, $\rewardFunc:\SAS \rightarrow \real$ is the reward function, and \discountFactor is the temporal reward discount factor. %
The objective is to find a policy \optimalPolicy such that:
$\optimalPolicy \in \argmax_{\policyFunc}\expectation[\returnFunc(\policyFunc)]$, where
$\returnFunc(\policyFunc) = \expectation_{\stateSym_{\timeSym},\stateSym_{\timeSym+1}\sim\transitionFunc;\actionSym_{\timeSym}\sim\policyFunc}\left[\sum_{\timeSym=0}^\infty\discountFactor^t \rewardFunc(\stateSym_t, \actionSym_t, \stateSym_{t+1})\right]$
is the expected return of policy \policyFunc.%

In this work, we focus on 
\textbf{transfer learning} (\TL), which is the improvement of learning a new task through the transfer of knowledge from a related task that has already been learned \citep{torrey_chapter_2009}. We model a collection of MDPs using a
\textbf{contextual MDP} (\CMDP) \citep{hallak_contextual_2015}, a 4-tuple \cmdpTup where \contextSpace is the context space, \stateSpace and \actionSpace are state and action spaces, and \mdpFunc is a mapping from a context $\contextSym\in\contextSpace$ to an MDP $\mdpFunc_\contextSym$ consisting of \stateSpace and \actionSpace but with distinct transition and reward functions, that is, $\mdpFunc_\contextSym=\mdpInContextTup$. We sometimes refer to context-induced MDPs as ``tasks'', and to the shared \stateSpace and \actionSpace as the ``environment''. In \cref{ex:house-contexts}, \contextSpace is the set of all house chores, \stateSpace and \actionSpace are the state of the house and the agent's capabilities, and \mdpFunc maps a chore \contextSym to an \MDP \mdpInContextSym that corresponds to completing the chore. 

\begin{figure*}[t]
\begin{minipage}[t][][t]{0.48\textwidth}
    \centering
    \includegraphics[width=1\textwidth]{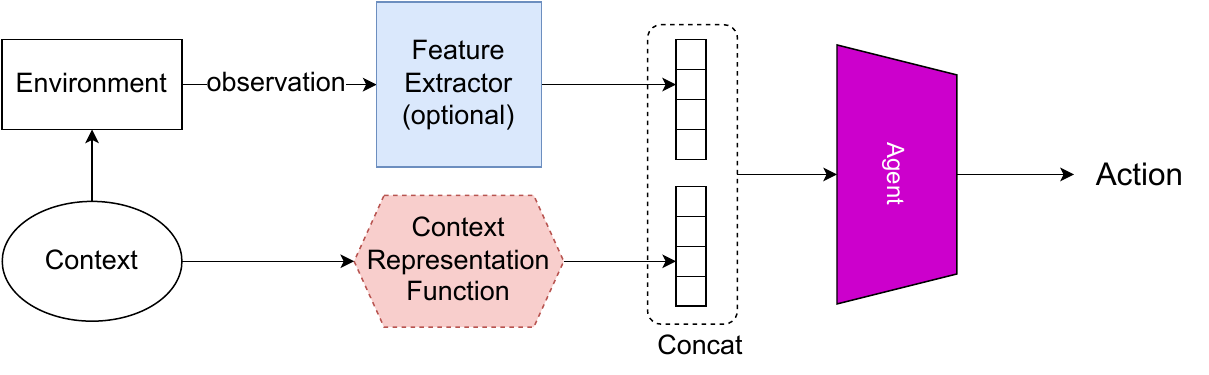}
    \subcaption{}
    \label{fig:cDRL-flow}
\end{minipage}\hfill
\begin{minipage}[t][][t]{0.48\textwidth}
    \centering
    \includegraphics[width=1\textwidth]{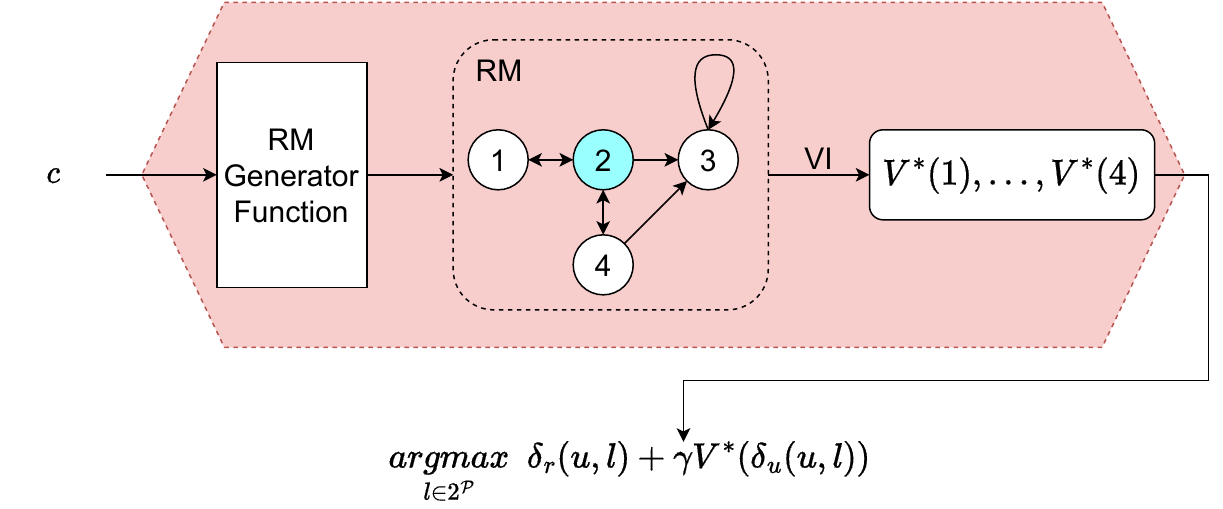}
    \subcaption{}
    \label{fig:c-prep}
\end{minipage}
\caption{(\subref{fig:cDRL-flow}) The general flow of transfer learning with a \CMDP. (\subref{fig:c-prep}) A visualization of the \cPrep context representation function. Context \contextSym is used to generate a task-specific \rmTerm.}
\end{figure*}

 \cref{fig:cDRL-flow} depicts the general flow of transfer learning over a \CMDP. The input is the observed MDP state and the current context. Optionally, the state representation is processed by a feature extractor to be represented as a vector. The context is represented via a {\em context representation function} that maps a context to a vector representation. The state representation is merged with the context representation (usually by concatenation), and the new representation is fed into a policy network that will determine the next action.

\citet{kirk_survey_2021} distinguish between two categories of context representations. The first type, known as {\em controllable context representations} (\ctxCTL), includes the necessary information to generate the MDP, which can be thought of as a transparent implementation of the environment generation process (implemented in \mdpFunc). The second type, {\em procedural content generation context representation} (\ctxPCG), conceals the MDP variables and only reveals information about the context identity, operating as a black box with no insight into the generation process.

Given a \CMDP \cmdpTup, transfer learning algorithms attempt to leverage knowledge from interactions with a set of {\em source contexts} $\srcContexts\subset\contextSpace$ to improve learning in a set of {\em target contexts} $\tgtContexts\subset\contextSpace$ such that $\srcContexts\cap\tgtContexts=\emptyset$. In \cref{ex:house-contexts}, \srcContexts is the set of contexts representing the chores it learns to do, including making coffee in a mug. Making coffee in a glass is a context in \tgtContexts. Policies learned after training in \srcContexts and \tgtContexts from scratch are the {\em source policy} and the {\em target policy}, respectively. The policy learned on \tgtContexts after training in \srcContexts is the {\em transferred policy}. Given a distribution \contextDistSym over \contextSpace, the objective is to optimize a chosen {\em transfer utility} \utilityFunc in expectation over sampled source and target context sets. Transfer utilities of interest in this work, suggested by \citet{taylor_transfer_2009}, are {\em jumpstart} (\js), {\em time to threshold} (\ttt), and {\em transfer ratio} (\tr). \js measures (zero-shot transfer) performance on the target contexts without additional training. \ttt measures the number of training steps taken until convergence to a policy of acceptable performance threshold (few-shot transfer). \tr measures the ratio of rewards accumulated over time by the agent using knowledge transfer against the agent that is trained from scratch, that is, how much the agent benefits from transfer (transfer relevance). Precise calculations are in \cref{app:eval}.

{\bf Reward machines} (\rmTerm{}s) \citep{toro_icarte_using_2018} are state machine abstractions of MDPs. Given a set of propositional symbols \rmPropSet, an \rmTerm is a 3-tuple $\rmSym = \rmTup$ where \rmAbsStateSpace is a set of abstract states, and $\rmAbsTransFunc:\rmAbsStateSpace\times2^{\rmPropSet}\rightarrow\rmAbsStateSpace$ and $\rmAbsRewardFunc:\rmAbsStateSpace\times2^{\rmPropSet}\rightarrow\real$ are the abstract transition and reward functions, respectively. Given the current abstract state $\rmAbsStateSym\in\rmAbsStateSpace$ and a subset of propositional symbols $\rmPropSym\subseteq\rmPropSet$ that hold true, $\rmAbsTransFunc(\rmAbsStateSym, \rmPropSym)$ is the next abstract state and $\rmAbsRewardFunc(\rmAbsStateSym, \rmPropSym)$ is the reward received for this transition. When $\rmAbsTransFunc(\rmAbsStateSym,\rmPropSym)=\rmAbsStateSym'$, \rmPropSym is called the abstract {\em \transitionLabelTerm} from \rmAbsStateSym to $\rmAbsStateSym'$. To connect between the abstraction and the underlying MDP, \rmPropSet is coupled with a {\em \stateLabelFuncTerm} $\stateLabelFunc:\SAS \rightarrow 2^{\rmPropSet}$ that maps state-action-state transitions in the MDP to abstract transition labels in the \rmTerm.

\begin{figure*}[t]
    \centering
    \begin{minipage}[b]{0.5\textwidth}
        \includegraphics[width=1\textwidth]{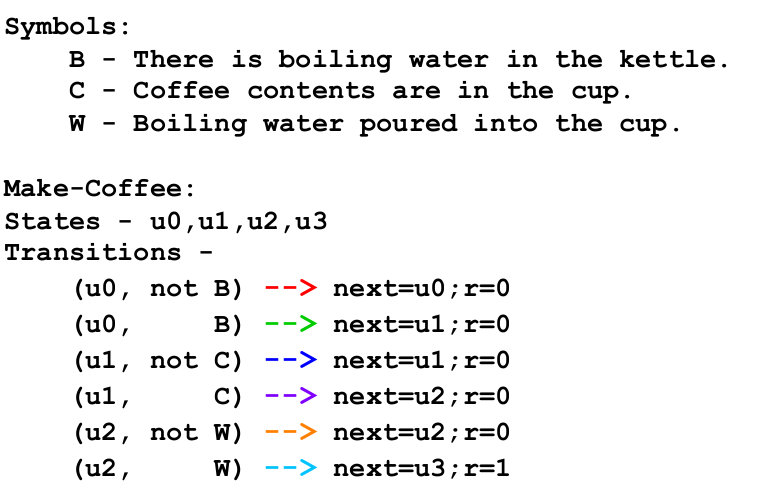}
        \subcaption{}
        \label{fig:example-text-rm}
    \end{minipage}\hfill
    \begin{minipage}[b]{0.46\textwidth}
        \includegraphics[width=1\textwidth]{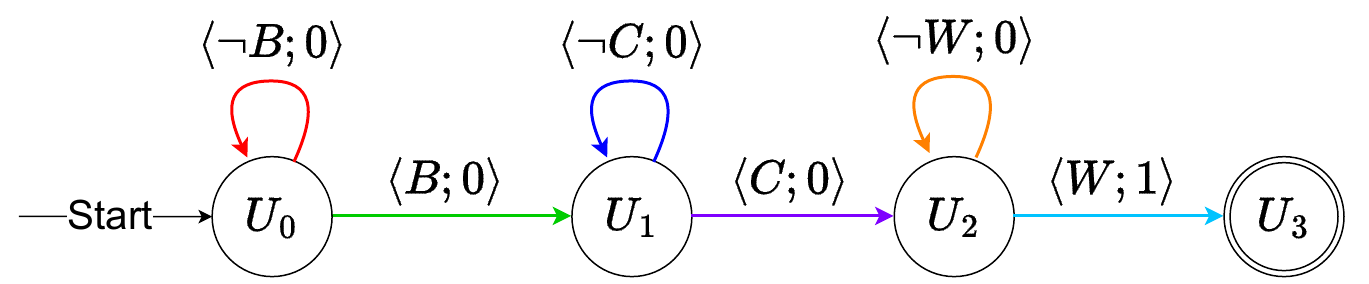}
        \subcaption{}
        \label{fig:example-rm}
        
        \includegraphics[width=1\textwidth]{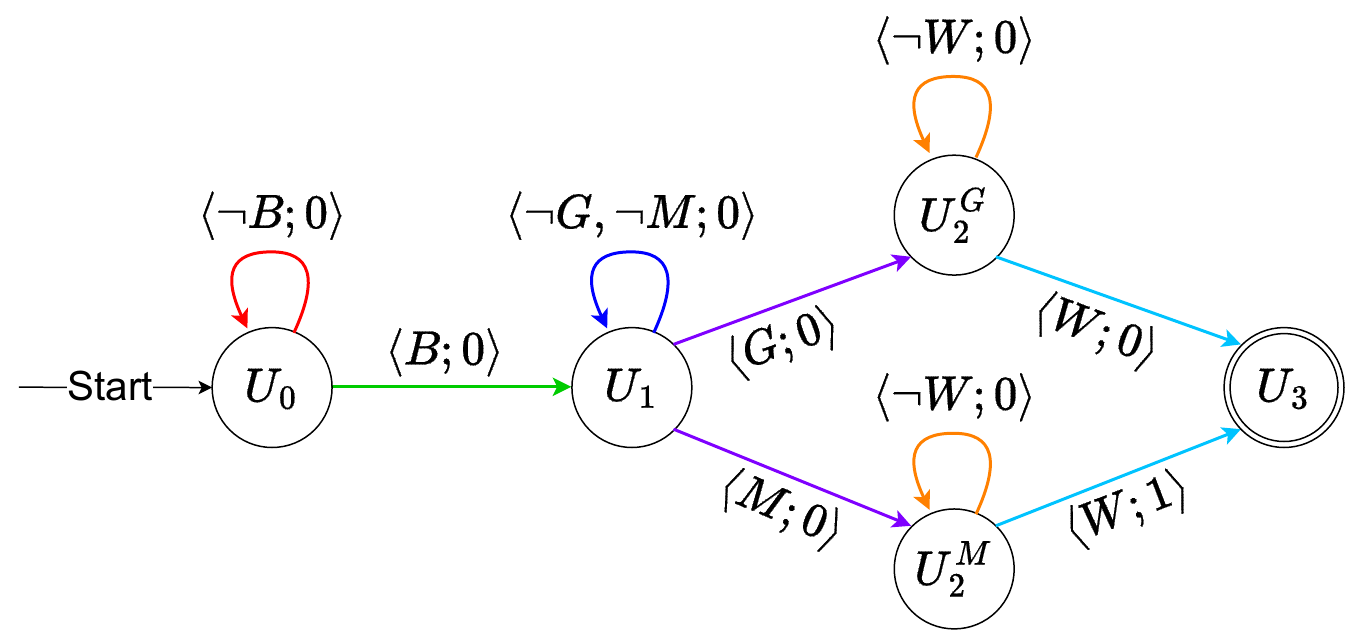}
        \subcaption{}
        \label{fig:example-rm-hi-res}
    \end{minipage}
    \cprotect\caption{(\subref{fig:example-text-rm}) A textual representation of the \rmTerm in \cref{ex:house-contexts} describing the \verb|Make-Coffee| task. (\subref{fig:example-rm}) A graph visualization of the textually defined \rmTerm. (\subref{fig:example-rm-hi-res}) An expansion of the \rmTerm that differentiates between mug and glass receptacles, described in \cref{sec:cprep}.}
    \label{fig:rm-graphs}
\end{figure*}

\cref{fig:example-text-rm} textually describes an \rmTerm for the task of making coffee in \cref{ex:house-contexts}. It defines abstract states $\rmAbsStateSym_0$ to $\rmAbsStateSym_3$ that each represents a high-level stage within the task of making a cup of coffee. The \rmTerm dictates that the agent must first boil some water, then put the coffee in the cup, and finally pour boiling water into the cup. These relationships are graphically visualized in \cref{fig:example-rm}.

The main benefits of \rmTerm{}s are that they represent transitions between abstract states using binary symbols that pertain to the state of the underlying MDP (through \stateLabelFunc) and provide dense rewards via reward shaping. As a result, an \rmTerm{} corresponding to some context divides its induced task into sub-tasks that each describe a stage in the process of solving the overall task, rewarding the agent upon completion of each sub-task. To employ sensible reward shaping, we use {\em potential-based reward shaping} \citep{ng_policy_1999} which, given a potential function \potentialFunc, defines a new abstract reward function $\rmAbsRewardFunc'(\rmAbsStateSym, \rmPropSym) = \rmAbsRewardFunc(\rmAbsStateSym, \rmPropSym) + \discountFactor\potentialFunc(\rmAbsTransFunc(\rmAbsStateSym, \rmPropSym)) - \potentialFunc(\rmAbsStateSym)$. \citet{toro_icarte_reward_2022-1} prove that potential-based reward shaping guarantees that optimal policies in an MDP for which rewards have been replaced with \rmTerm rewards are optimal using the \rmTerm reshaped rewards. Moreover, it is empirically shown that using \rmTerm reshaped rewards can significantly expedite policy convergence for RL agents.

\section{Contextual Pre-Planning (\cPrep) for Transfer Learning}\label{sec:cprep}

We aim to improve transfer in multi-task domains modeled as \CMDP{}s. \citet{benjamins_contextualize_2022} proved that the policy must be conditioned on the context itself to guarantee optimality. Therefore, it is crucial to represent the context such that the agent can generalize across contexts. For this, we use \rmTerm{}s to represent contexts and offer a novel way to enhance the agent's ability to exploit its previous experiences in new settings. Since our focus is on exploiting the structure of the \rmTerm{}s for transfer and not on their generation, we assume that the \rmTerm generator function is given as input, which can be based on domain knowledge, learned from demonstration \citep{camacho_reward_2021}, or learned via discrete optimization \citep{toro_icarte_learning_2019}.

\citet{camacho_reward_2021} exploited \rmTerm{}s to expedite learning in single-task domains by providing the agent with the current abstract state. We instead focus on transfer learning and provide the next desired abstract transition from the current \rmTerm abstract state as contextual input at each timestep. Essentially, we guide the agent through optimal paths in the \rmTerm with abstract transitions, represented using a set of symbols that is shared across all tasks. Upon transfer, the agent can expedite transfer learning by exploiting abstract transitions and leveraging prior knowledge of encountered symbols in the new task.
This may be beneficial for learning in general but is key in transfer settings as it provides reusable representations between tasks. 

\paragraph{\cPrep Context Representation Function.}

Based on the above intuition, we propose {\bf \em \cPrepTerm} (\cPrep) for leveraging information in context-specific \rmTerm{}s\footnote{Code: \url{https://github.com/CLAIR-LAB-TECHNION/C-PREP}}. %
For each task, \cPrep generates an \rmTerm \rmTup with abstract transitions represented using a shared symbol set, i.e., all \rmTerm are defined using the symbol set \rmPropSet. Using {\em value iteration} (VI) \citep{bellman_markovian_1957}, we find an optimal policy in the \rmTerm as if it were a deterministic \MDP. We note that VI is one of many possible choices, but we use it here to compare to previous methods. We then give the agent an optimal abstract transition label in the \rmTerm from the current abstract state \rmAbsStateSym (as dictated by the \rmTerm policy), i.e., a transition label $\rmPropSym$ such that $\rmAbsTransFunc(\rmAbsStateSym,\rmPropSym)$ is the next state on a (discounted) reward-maximizing path in the \rmTerm. Intuitively, we wish to guide the agent towards an optimal path within the \rmTerm. 

\cPrep relies on providing the next desired abstract transition in the \rmTerm to the agent. However, since there is no direct representation of actions in the \rmTerm{}, standard VI does not apply. We, therefore, use a variant of VI, as suggested by \citet{toro_icarte_reward_2022-1}, with the following update rule over the abstract states of \rmTerm \rmSym.
 \begin{equation}\label{eq:rm-vi}
    \valueFunc_{\rmSym}^{\viIndex}(\rmAbsStateSym) = \max_{\rmPropSym\in 2^{\rmPropSet}}\left[\rmAbsRewardFunc(\rmAbsStateSym,\rmPropSym) + \discountFactor \valueFunc_{\rmSym}^{\viIndex - 1}(\rmAbsTransFunc(\rmAbsStateSym,\rmPropSym))\right]
\end{equation}
where $\valueFunc_{\rmSym}^{\viIndex}$ is the value of abstract state \rmAbsStateSym at iteration \viIndex ($\valueFunc_{\rmSym}^{0}=0$), and $\rmAbsTransFunc(\rmAbsStateSym,\rmPropSym)$ and $\rmAbsRewardFunc(\rmAbsStateSym,\rmPropSym)$ are the next abstract state and reward received for achieving transition label \rmPropSym at abstract state \rmAbsStateSym, respectively.
To show the relationship between this rule and VI for \MDP{}s, we define $\mdpSym_{\rmSym} = \tuple{\rmAbsStateSpace,2^{\rmPropSet},\transitionFunc,\rewardFunc,\discountFactor}$ where $\transitionFunc(\rmAbsStateSym,\rmPropSym,\rmAbsTransFunc(\rmAbsStateSym,\rmPropSym)) = 1$ and $\rewardFunc(\rmAbsStateSym,\rmPropSym,\rmAbsStateSym') = \rmAbsRewardFunc(\rmAbsStateSym,\rmPropSym)$. We observe that the VI update rule for $\mdpSym_{\rmSym}$, denoted $\valueFunc^{\viIndex}$, is equivalent to $\valueFunc_{\rmSym}^{\viIndex}$. Formally, 
\begin{equation*}
    \begin{split}
        \valueFunc^{\viIndex}(\rmAbsStateSym) = & \max_{\rmPropSym\in2^\rmPropSet}\sum_{\rmAbsStateSym' \in \rmAbsStateSpace}\transitionFunc(\rmAbsStateSym,\rmPropSym,\rmAbsStateSym')(\rewardFunc(\rmAbsStateSym,\rmPropSym,\rmAbsStateSym') + \discountFactor \valueFunc^{\viIndex - 1}(\rmAbsStateSym')) \\
        = & \max_{\rmPropSym\in2^\rmPropSet}\rewardFunc(\rmAbsStateSym,\rmPropSym,\rmAbsTransFunc(\rmAbsStateSym,\rmPropSym)) + \discountFactor \valueFunc^{\viIndex - 1}(\rmAbsTransFunc(\rmAbsStateSym,\rmPropSym)) \\
        = & \max_{\rmPropSym\in2^\rmPropSet}\rmAbsRewardFunc(\rmAbsStateSym,\rmPropSym) + \discountFactor \valueFunc^{\viIndex - 1}(\rmAbsTransFunc(\rmAbsStateSym,\rmPropSym))) = \valueFunc_{\rmSym}^{\viIndex}(\rmAbsStateSym)
    \end{split}
\end{equation*}
Thus, to identify optimal abstract transitions, we can find an abstract optimal policy in \rmTerm \rmSym by
using VI to find an optimal policy \optimalPolicy in $\mdpSym_{\rmSym}$.

Given the current abstract state \rmAbsStateSym, from which there may be multiple optimal abstract transitions, \cPrep samples an optimal abstract transition \rmPropSym from $\optimalPolicy(\cdot|\rmAbsStateSym)$. Since \optimalPolicy is optimal in deterministic \MDP $\mdpSym_{\rmSym}$:
\begin{equation*}
    supp(\optimalPolicy(\cdot|\rmAbsStateSym)) \subset \argmax_{\rmPropSym\in 2^{\rmPropSet}}\left(\rmAbsRewardFunc(\rmAbsStateSym,\rmPropSym) + \discountFactor \valueFunc^*(\rmAbsTransFunc(\rmAbsStateSym,\rmPropSym))\right)
\end{equation*}
where $supp(\optimalPolicy(\cdot|\rmAbsStateSym))$ is the support set of probability distribution $\optimalPolicy(\cdot|\rmAbsStateSym)$. Thus, any transition we sample from \optimalPolicy is one that maximizes discounted return in the \rmTerm.

Based on the above formulations, the \cPrep context representation function (depicted in \cref{fig:c-prep}) operates in a three-step process: (1) generate an \rmTerm $\rmSym = \rmGenFunc(\mdpInContextSym)$ for the current context \contextSym, (2) find an optimal policy \optimalPolicy in $\mdpSym_{\rmSym}$, (3) at each timestep, sample an optimal transition $\rmPropSym\sim\optimalPolicy(\cdot,\rmAbsStateSym)$ given the current \rmTerm abstract state \rmAbsStateSym and return it.

Throughout training, the \cPrep \rmTerm~generation function updates its returned representation according to the current abstract state. To notify the agent that a correct (or incorrect) abstract transition has been completed, we provide additional rewards that emphasize the executed abstract transition's quality. For this, we employ potential-based reward-shaping as defined in Section \ref{sec:bg}. As it is already calculated, we use $\valueFunc^*$ as the potential function \potentialFunc to generate the reward signal that is provided to the agent instead of the original MDP reward. In the \rmTerm described in \cref{fig:example-text-rm}, the agent will receive a higher reward for transitioning from state $\rmAbsStateSym_0$ to $\rmAbsStateSym_1$ rather than loop back to itself because this brings it closer to the abstract goal state.

\paragraph{Transfer Learning with \cPrep}

The input to our setting includes a \CMDP $\cmdpTup$ and an {\em \rmGenFuncTerm} \rmGenFunc that maps each context-induced task \mdpInContextSym to its corresponding \rmTerm $\rmGenFunc(\mdpInContextSym) = \rmInContextTup$ which is defined over shared symbol set \rmPropSet. 

The \cPrep context representation function can be integrated into any algorithm following the transfer learning flow depicted in \cref{fig:cDRL-flow}. \cref{alg:train} (\cref{app:alg}) demonstrates an implementation of a \algDQN \citep{mnih_human-level_2015} for transfer learning settings using \cPrep as the context representation function and \rmTerm reward shaping. The key differences between this implementation and the standard \algDQN are that the algorithm initially generates an \rmTerm for the sampled context, calculates its state values, and reshapes the \rmTerm rewards. States encountered in the episode are augmented by the \cPrep context representation according to the \rmTerm transition. Rewards are replaced with the reshaped rewards from the \rmTerm according to the achieved abstract transition at that timestep.

We note that the ability of
\cPrep to support transfer depends on the {\em resolution} of the generated \rmTerm{}s, i.e., how well the generated \rmTerm{}s represent the context space. If the set of propositional symbols \rmPropSet is too abstract, the generated \rmTerm{}s do not sufficiently distinguish between contexts. In contrast, if it is too refined, computation time may increase due to running VI in huge tables for every context.

In \cref{ex:house-contexts}, when training to make coffee in a mug, the agent learns to pour water into the mug and should exploit this capability upon transferring to the task of making coffee in a glass.
\cref{fig:rm-graphs} shows two different \rmTerm{}s that can be used to describe this setting. The \rmTerm in \cref{fig:example-rm} does not differentiate between a mug and a glass, as they are both encapsulated by the ``cup'' symbol \verb|C|. In contrast, the \rmTerm in \cref{fig:example-rm-hi-res} distinguishes between the tasks of making coffee in a mug and in a glass, rewarding the agent only for the former (when transitioning from $\rmAbsStateSym^M_2$ to $\rmAbsStateSym_3$).  
Including both mug and glass events demonstrates a case with two possible ways to perform a task, differentiated by the reward function.
\section{Empirical Evaluation}\label{sec:eval}
The objective of our empirical evaluation is to examine whether agents using \cPrep exhibit improved performance on transfer utilities of interest.

\subsection{Experimental Setup}\label{sec:eval:envs}

\paragraph{Environments:}
We test our method in four environments with compound and long-horizon tasks and sparse reward signals\footnote{Our code base is described in \cref{app:multi-taxi}.}:

\noindent{\bf Grid Navigation (GN):} An agent must reach a specified destination on a grid. The state space consists of the agent's current location and the action space includes moving in one of the four cardinal directions and a "done" action to be called upon arrival at the destination. \\
{\bf Multi Points-of-Interest (MP):} The agent navigates to multiple destinations {\em in any order}. The state space consists of the agent's location and an indicator of whether a certain destination has already been visited. The action space is as in GN but with an "arrived" action replacing "done". \\
{\bf Pick-Up and Drop-Off (PD):} An agent picks up and drops off passengers at their destinations. The state space is as in MP with indicators for passengers that have been dropped off at their destinations. In addition to navigation actions, the action space contains "pick-up" and "drop-off" actions. \\
{\bf Ordered Navigation (ON):} The agent must navigate to specified destinations {\em in a specific order}. The state and action spaces are as in MP.

All maps are $6\times6$. At every timestep, the agent receives a reward of 1 when achieving the environment objective and 0 otherwise, and the discount factor is $\discountFactor = 0.99$.

\paragraph{Defining the \CMDP Spaces:}
The environments described above include pairs of state and action spaces. To define a \CMDP we couple them with the following context spaces:
\noindent{\bf Entity Location (EL):} The context indicates the locations of core entities in the environment, e.g., passenger locations and drop-off destinations.\\
{\bf Changing Map (CM):} The context indicates the number and location of walls in the grid. \\
{\bf Point-of-Interest Order (PO):} The context indicates the order of the locations to visit.\\
Each GN, MP, and PD environment is used with both the EL and CM context spaces. The ON environment is paired with the PO context space. Contexts are represented using \ctxCTL representations (see \cref{sec:bg}). For full details on \ctxCTL context representations, see \cref{app:ctx-reps}.

\paragraph{Transfer Session:}
Each training session begins by randomly sampling two disjoint context sets from the \CMDP{}s described above; the source set \srcContexts and the target set \tgtContexts. We adopt ``training protocol {\bf B}'' of \citet{kirk_survey_2021} such that the size of $\srcContexts$ is much smaller than the size of the context space. The agent initially trains on tasks induced by \srcContexts for \nSrc steps and then continues its training in \tgtContexts for additional \nTgt steps. We record performance progress during and after training. 
For full details see \cref{app:training}.

\paragraph{Context Representations:}
We vary the \rmTerm information exposed to the agent, using the following representations:

\begin{itemize}[leftmargin=*]
    \item \ctxCTL: controllable context representation without \rmTerm{}s (same baseline in \citep{toro_icarte_using_2018}).
    \item \ctxCTL{}+\modRS \citep{toro_icarte_using_2018}: adds dense reshaped \rmTerm rewards to the current context's reward functions.
    \item \ctxCTL{}+\modLTL{}+\modRS \citep{camacho_reward_2021}: adds the {\em Last Transition Label} (\modLTL) as an additional context representation that is the current assignment of symbols in \rmPropSet.
    \item \ctxCTL{}+\cPrep (\textbf{ours}): adds the \cPrep context representation: {\em Desired Transition Label} (\modDTL), with \modRS.
    \item \cPrep (\textbf{ours}): \cPrep context representation without a \ctxCTL context representation
\end{itemize}

In \cref{app:pcg} we show additional experiments using \ctxPCG context representations in lieu of \ctxCTL.

\paragraph{Reported Metrics:}

During each training session, we evaluate the source, target, and transferred policies on the context set on which it is trained at $100$ uniformly spaced evaluation points. At each evaluation point, we record the policy's average return on 50 sampled contexts. Each training session is repeated $5$ times, using different random seeds. From the computed average returns, we calculate the transfer utilities defined in \cref{sec:bg}: \js, \ttt, and \tr (see \cref{app:eval} for the formula used to compute these measures). We aggregate the results using interquartile mean (IQM) and calculate the standard deviation and stratified bootstrap 95\% confidence intervals \citep{agarwal_deep_2021}. To report results for different performance thresholds, we plot the \ttt as a function of the threshold. We measure the IQM area under the curve (\auc) of this function, denoted $\ttt_{\auc}$.

\subsection{Results}\label{sec:eval:res}

First, to examine the performance over the entire transfer session,  \cref{tab:main} shows the interquartile mean (IQM) and standard deviation of the measured transfer utilities ($\text{\ttt}_{\auc}$, \js, \tr) for all tested configurations using a \ctxCTL context representation. The best results for each \CMDP (row) are marked in bold. Negative \tr values that indicate non-beneficial transfer are italicized. 

Our method performed best in terms of $\ttt_{\auc}$ and \js in all but two \CMDP{}s: (1) in GN (shortest horizon) with both context spaces (EL and CM), \ctxCTL{}+\modLTL{}+\modRS performs best in terms of $\ttt_{\auc}$; (2) in GN with EL context space, using \ctxCTL alone performs best in terms of \js (note that GN with the EL context space has a small context space providing less data for transfer; see \cref{app:training}). Notably, in PD, which is the longest horizon task, our method outperforms all other configurations. Compared to the highest performing baseline, we see $\ttt_{\auc}$ improvements of 22.84\% in the context space CM and 42.31\% in the context space EL, and \js improvements of 11.86\% in CM and 36.5\% in EL. \tr results show low and negative \tr values for most configurations. Our method is the only one with positive \tr throughout all tasks. In the PO environment (longest horizon), we see a performance improvement of over 300\% when using \cPrep compared to the next best configuration.

\begin{table*}[t]
    \centering
    \resizebox{1\textwidth}{!}{
    \bgroup
    \def\arraystretch{1.2}
    \begin{tabular}{|c|c|c?c|c|c|c|c|c|c|}
\hline
 Utility   & Context Space   & Environment   & CTL           & CTL+RS        & CTL+LTL+RS    & CTL+C-PREP (ours)    & C-PREP (ours)        \\
\Xhline{4\arrayrulewidth}
\parbox[t]{2mm}{\multirow{7}{*}{\rotatebox[origin=c]{90}{$\ttt_{\auc}$}}}   & \multirow{3}{*}{EL} & GN   & 25.41 $\pm$ 7.27  & 6.37 $\pm$ 0.71   & \textbf{6.21 $\pm$ 0.42}  & 6.42 $\pm$ 0.44    & 42.15 $\pm$ 3.95  \\
        &   & MP   & 94.77 $\pm$ 6.02  & 44.78 $\pm$ 12.06 & 19.88 $\pm$ 9.36  & \textbf{18.32 $\pm$ 9.27}  & 86.43 $\pm$ 1.89  \\
        &   & PD   & 97.39 $\pm$ 1.57  & 95.18 $\pm$ 3.98  & 37.58 $\pm$ 23.38 & \textbf{21.68 $\pm$ 6.55}  & 88.35 $\pm$ 1.59  \\
        \cline{2-8}
        & \multirow{3}{*}{CM}    & GN   & 42.18 $\pm$ 1.58  & 16.30 $\pm$ 2.35  & \textbf{7.14 $\pm$ 1.41}  & 7.64 $\pm$ 1.72    & 32.98 $\pm$ 0.68  \\
        &     & MP   & 71.90 $\pm$ 10.62 & 26.48 $\pm$ 15.51 & 14.24 $\pm$ 8.85  & \textbf{13.64 $\pm$ 8.60}  & 47.74 $\pm$ 11.46 \\
        &     & PD   & 86.30 $\pm$ 18.81 & 54.19 $\pm$ 21.64 & 37.52 $\pm$ 25.46 & \textbf{28.95 $\pm$ 24.15} & 51.02 $\pm$ 15.46 \\
        \cline{2-8}
        & PO    & ON   & 98.04 $\pm$ 0.00  & 95.15 $\pm$ 7.66  & 97.50 $\pm$ 5.59  & 32.93 $\pm$ 10.21  & \textbf{22.54 $\pm$ 1.57} \\
 \Xhline{4\arrayrulewidth}
 \parbox[t]{2mm}{\multirow{7}{*}{\rotatebox[origin=c]{90}{\js}}}   & \multirow{3}{*}{EL} & GN   & \textbf{0.28 $\pm$ 0.11} & 0.06 $\pm$ 0.06  & 0.07 $\pm$ 0.09  & 0.05 $\pm$ 0.05  & 0.01 $\pm$ 0.05  \\
         &   & MP   & 0.03 $\pm$ 0.06  & 0.26 $\pm$ 0.15  & \textbf{0.48 $\pm$ 0.26}  & \textbf{0.49 $\pm$ 0.17} & 0.01 $\pm$ 0.02  \\
         &   & PD   & 0.02 $\pm$ 0.02  & 0.03 $\pm$ 0.01  & 0.34 $\pm$ 0.20  & \textbf{0.38 $\pm$ 0.12} & 0.00 $\pm$ 0.01  \\
         \cline{2-8}
         & \multirow{3}{*}{CM}    & GN   & 0.42 $\pm$ 0.03  & 0.55 $\pm$ 0.06  & 0.73 $\pm$ 0.05  & \textbf{0.75 $\pm$ 0.08} & 0.49 $\pm$ 0.06  \\
         &    & MP   & 0.23 $\pm$ 0.09  & 0.49 $\pm$ 0.14  & 0.66 $\pm$ 0.11  & \textbf{0.68 $\pm$ 0.10} & 0.42 $\pm$ 0.10  \\
         &     & PD   & 0.08 $\pm$ 0.19  & 0.28 $\pm$ 0.19  & 0.38 $\pm$ 0.27  & \textbf{0.52 $\pm$ 0.25} & 0.31 $\pm$ 0.19  \\
         \cline{2-8}
         & PO    & ON   & 0.00 $\pm$ 0.00  & 0.00 $\pm$ 0.00  & 0.00 $\pm$ 0.00  & 0.14 $\pm$ 0.26  & \textbf{0.75 $\pm$ 0.02} \\
\Xhline{4\arrayrulewidth}
\parbox[t]{2mm}{\multirow{7}{*}{\rotatebox[origin=c]{90}{\tr}}}   & \multirow{3}{*}{EL} & GN   & \textit{-0.11 $\pm$ 0.10} & \textbf{0.13 $\pm$ 0.04} & 0.08 $\pm$ 0.03  & 0.07 $\pm$ 0.03  & 0.04 $\pm$ 0.02  \\
         &   & MP   & \textit{-0.99 $\pm$ 0.01} & \textbf{0.24 $\pm$ 0.08} & 0.24 $\pm$ 0.18  & 0.16 $\pm$ 0.12  & 0.06 $\pm$ 0.11  \\
         &   & PD   & \textit{-1.00 $\pm$ 0.00} & \textit{-0.99 $\pm$ 0.21} & \textit{-0.14 $\pm$ 0.38} & \textbf{0.14 $\pm$ 0.09} & \textit{-0.05 $\pm$ 0.12} \\
         \cline{2-8}
         & \multirow{3}{*}{CM}    & GN   & \textit{-0.11 $\pm$ 0.04} & 0.08 $\pm$ 0.04  & \textbf{0.11 $\pm$ 0.01} & \textbf{0.11 $\pm$ 0.01}  & 0.08 $\pm$ 0.04  \\
         &     &MP   & \textit{-0.85 $\pm$ 0.32} & \textbf{0.29 $\pm$ 0.13} & 0.26 $\pm$ 0.06  & 0.21 $\pm$ 0.06  & 0.07 $\pm$ 0.05  \\
         &     & PD   & \textit{-0.94 $\pm$ 0.25} & \textit{-0.06 $\pm$ 0.29} & \textit{-0.05 $\pm$ 0.20} & 0.06 $\pm$ 0.19  & \textbf{0.07 $\pm$ 0.07} \\
         \cline{2-8}
         & PO    & ON   & 0.00 $\pm$ 0.00  & \textbf{2.04 $\pm$ 6.94}\tablefootnote{Target policy scored close to 0 on all seeds.} & \textit{-0.87 $\pm$ 0.90} & 0.69 $\pm$ 0.20  & 0.31 $\pm$ 0.03  \\
\hline
\end{tabular}
\egroup
}
\caption{IQM and standard deviation transfer utilities of configurations with \ctxCTL. Bold cells indicate best performance in each row. Italic cells indicate negative TR}
    \label{tab:main}
\end{table*}

\begin{figure*}[t]
    \centering
    \begin{minipage}[t][][t]{0.65\textwidth}
        \includegraphics[width=1\textwidth]{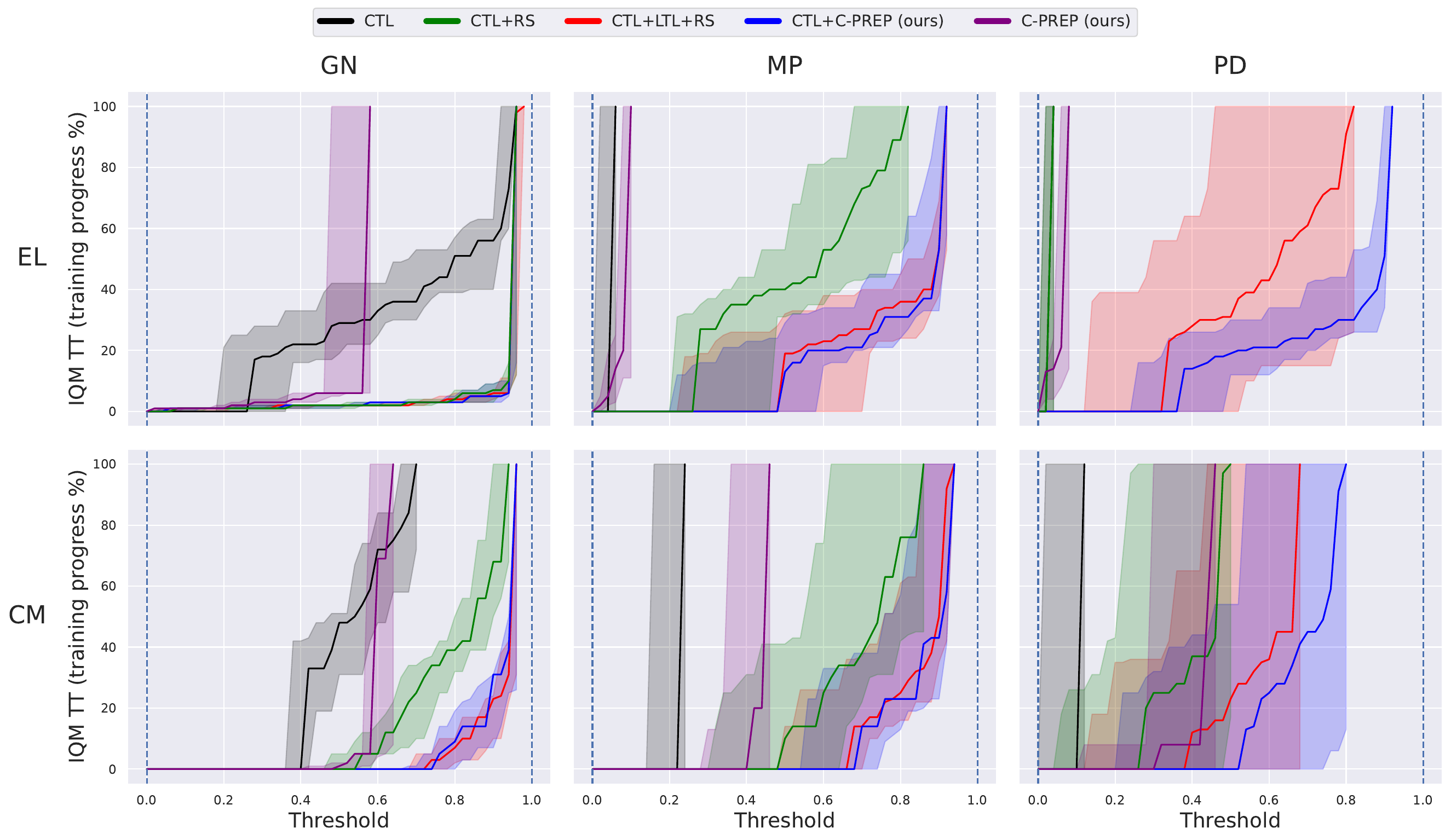}
        \subcaption{GN, MP, and PD environments with EL and CM context spaces}
        \label{fig:multi-ttt:main}
    \end{minipage}\hfill
    \begin{minipage}[t][][t]{0.33\textwidth}
        \includegraphics[width=1\textwidth]{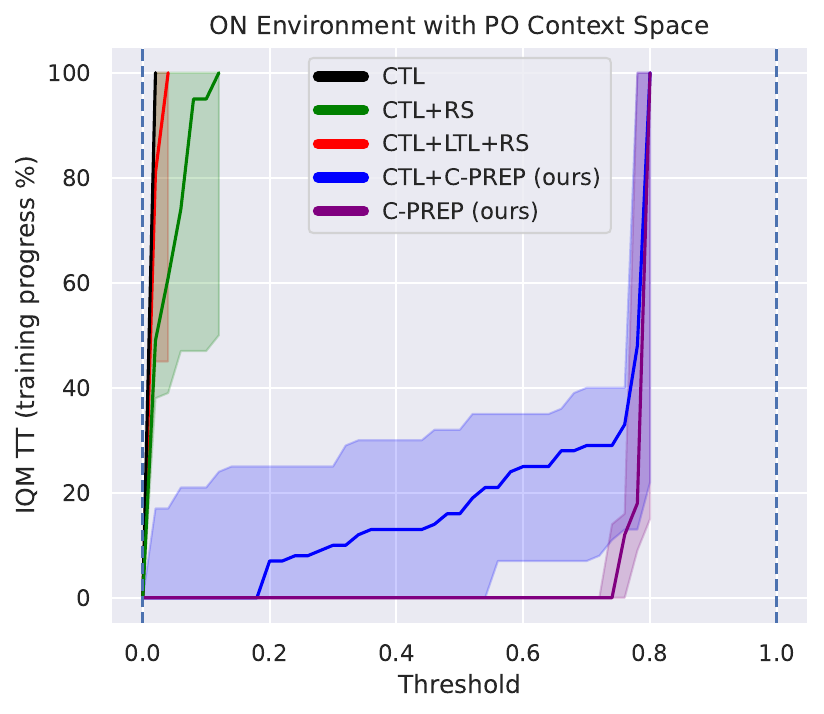}
        \subcaption{ON task with PO context space}
        \label{fig:res:on-ttt}
    \end{minipage}
    \caption{The IQM \ttt of configurations using \ctxCTL as a function of the threshold (lower is better).}
    \label{fig:multi-ttt}
\end{figure*}

Next, we examine the achieved threshold performance throughout training. \cref{fig:multi-ttt} visualizes the IQM \ttt results, measured in training progress (percentage) as a function of the threshold, i.e., the curve from which we derive $\ttt_{\auc}$. The shaded areas are stratified bootstrap 95\% confidence intervals. Each row corresponds to a context space. Each column corresponds to an environment.  Agents using \modRS in environments tested with the CM and EL context show similar performance in GN, but a performance gap (in favor of our method) widens as the task horizon grows. In the PO environment, our method is the only one that can be seen converging to a high-performing policy.

\cref{app:pcg} shows results of experiments using uninformative \ctxPCG representations in place of \ctxCTL. \ttt and \js results are similar to those presented with \ctxCTL representations. \tr results show that for all configurations, it is non-beneficial to use \ctxPCG representations for transfer due to severe overfitting. \cref{app:abl} presents ablation results that show that all components of \cPrep are required to achieve the best results. \cref{app:add-res} shows results for additional experiments on sample efficiency (in terms of number of contexts) and generalization capabilities of \cPrep using \ctxPCG.

\subsection{Discussion}\label{sec:eval:disc}

Results demonstrate that \cPrep improves transfer performance in more complex tasks
without hindering performance on simpler tasks. As visualized in \cref{fig:multi-ttt:main}, all methods perform similarly in the GN environment (short horizon), but \cPrep opens a performance gap in \ttt that increases with the difficulty of the environment. The \tr results show that only our method is beneficial for transfer in all tasks, as is evidenced by the negative \tr values reported for all other configurations. In the PO environment, only agents using \cPrep achieve a threshold greater than 0.2. Furthermore, since the \rmTerm in this case differentiates all tasks, it is preferable to use \cPrep without \ctxCTL. We observe that the \js performance is approximately 93\% of the maximum achieved performance threshold, which is reached in less than 20\% of the training progress.

We examine the performance of \cPrep using partial resolution \rmTerm{}s, i.e., some tasks may be represented with the same \rmTerm. For this, we use \cPrep alone. In the GN, MP, and PD environments, the agent will achieve a threshold performance of no more than 50\% of \cPrep's performance {\it with} \ctxCTL. \cref{fig:multi-ttt:main} shows that \cPrep without \ctxCTL achieves medium to low performance depending on the environment and context space. We attribute this to the low coverage of tasks with the partial \rmTerm resolution. These \rmTerm{}s (detailed in \cref{app:rm}) cover approximately 25\% of the tasks in GN and approximately 6\% of the MP and PD tasks. We conclude that \cPrep with partial resolution \rmTerm{}s cannot compensate for missing contextual information. However,  \cref{fig:res:on-ttt} shows that in the ON environment, where the \rmTerm{}s are of full resolution, it is preferable to use \cPrep without additional context. We hypothesize that this is due to the large overlap in contextual data between the \cPrep context representation and \ctxCTL, making the information in \ctxCTL irrelevant. This use of local context illustrates the advantage of solving the context as a series of smaller, simpler contexts.

We show additional results in experiments using \ctxPCG in place of \ctxCTL in \cref{app:pcg} to examine the case of uninformative global context representations. Here, we notice that in 60\% of the runs, it was more beneficial to train from scratch in \tgtContexts than to transfer from \srcContexts. \cref{fig:res:tsf-pcg} in the appendix visualizes \ttt performance of \ctxPCG configurations and shows hindered performance compared to those in \cref{fig:multi-ttt:main} that use \ctxCTL representations.

Ablation results (found in \cref{app:abl}) show the importance of every component of \cPrep. We notice that both $\ttt_{\auc}$ and \js can be improved by up to 12\% in five out of seven tasks by adding the \modLTL{} modification to \ctxCTL{}+\cPrep in the most complex task. We hypothesize that this will yield an even greater benefit in scenarios where it is harder to infer the current abstraction label.

Additional results in \cref{app:add-res} reveal two interesting capabilities of \cPrep. First, \cPrep is more sample efficient in terms of the number of source contexts it is trained on, that is, \cPrep needs to train on fewer source contexts to achieve similar or better transfer performance than other tested configurations. Second, adding \rmTerm information when using \ctxPCG significantly improves generalization capabilities at the beginning of training. We see a spike in performance in the first 1M training steps, hinting at the potential of using \rmTerm{}s for learning generalized state representations.

\section{Related Work}\label{sec:related}

DRL agents are susceptible to overfitting to the context in which they were trained. \citet{leike_ai_2017} show that small changes to a single detail or obstacle could result in extreme performance degradation. \citet{danesh_re-understanding_2021} demonstrate that simple RL agents overfit to the training settings such that they completely ignore observations. One solution is to train the agent on a distribution of contexts, rather than a single one \citep{zhang_study_2018}. However, once the context distribution departs from the training distribution, the performance drops despite the knowledge obtained during training \citep{agarwal_contrastive_2020}. We focus on transferring knowledge to expedite training in novel tasks. 

There are several approaches to improve transfer learning in DRL. Meta-learning methods \citep{finn_model-agnostic_2017,wang_learning_2017,duan_rl2_2016} ``learn to learn'', thereby expediting the agent's adaptation to new surroundings. Model-based methods \citep{shrestha_deepaveragers_2020, tamar_value_2016} learn an approximate model of the world, where agents can plan for different contexts using the same model. Bayes-adaptive exploration \citep{dorfman_offline_2021, zintgraf_varibad_2019} learn how to best explore new environments. Other methods include disentangling latent representations in various ways to improve adaptation \cite{dunion_conditional_2023,dunion_temporal_2023}. All of the above rely on additional exploration to determine the context before learning to solve it. In contrast, we use contextual information to understand the task a priori to reduce exploration. Using \CMDP{}s, we view the context as additional input to the agent \citep{langford_contextual_2017,hallak_contextual_2015}. 

To improve few-shot transfer, our method represents contexts as \rmTerm{}s \citep{toro_icarte_using_2018}. Previous work shows that \rmTerm{}s can be used to expedite learning of a single fully observable or partially observable context \citep{camacho_reward_2021,vaezipoor_ltl2action_2021}. The latter work uses {\em linear temporal logic}, which uses structures similar to reward machines with time-related constraints. Note that this is abbreviated \modLTL, not to be confused with the {\em last transition label} in this paper. We utilize \rmTerm{}s to represent contextual information, resulting in better sample efficiency and few-shot transfer learning capabilities in multi-context settings.

\section{Conclusion}
We presented  {\bf \em \cPrepTerm} (\cPrep) as a novel context representation function and showed how it enhances zero-shot and few-shot transfer for DRL agents. 
\cPrep exploits \rmTerm{}s by planning on them and providing the agent with a representation of the next desired transition. Our empirical evaluation demonstrates \cPrep's ability to improve sample efficiency and different transfer utilities, especially for tasks of increasing difficulty.

To focus on improving transfer using \rmTerm representations, we assumed the \rmTerm{} generation function is given as input. Future work will include a theoretical analysis of the conditions under which a context representation is guaranteed to enhance transfer and the development of methods for learning to generate appropriate \rmTerm{}s. As a second extension, we intend to examine alternative symbolic representations beyond \rmTerm{}s for enhancing learning and transfer, as well as consider our context representation function's effect in setting in which {\em options} \citep{sutton_between_1999-1,illanes_symbolic_2020}
 are used to distinguish between sub-tasks. Finally, we plan to examine our representation in real-world settings in which transfer may be beneficial. 

\bibliography{references}

\begin{thebibliography}{39}
\providecommand{\natexlab}[1]{#1}

\bibitem[{Agarwal et~al.(2020)Agarwal, Machado, Castro, and Bellemare}]{agarwal_contrastive_2020}
Agarwal, R.; Machado, M.~C.; Castro, P.~S.; and Bellemare, M.~G. 2020.
\newblock Contrastive {Behavioral} {Similarity} {Embeddings} for {Generalization} in {Reinforcement} {Learning}.
\newblock In \emph{International {Conference} on {Learning} {Representations}}.

\bibitem[{Agarwal et~al.(2021)Agarwal, Schwarzer, Castro, Courville, and Bellemare}]{agarwal_deep_2021}
Agarwal, R.; Schwarzer, M.; Castro, P.~S.; Courville, A.~C.; and Bellemare, M. 2021.
\newblock Deep {Reinforcement} {Learning} at the {Edge} of the {Statistical} {Precipice}.
\newblock In \emph{Advances in {Neural} {Information} {Processing} {Systems}}, volume~34, 29304--29320. Curran Associates, Inc.

\bibitem[{Bellman(1957)}]{bellman_markovian_1957}
Bellman, R. 1957.
\newblock A {Markovian} {Decision} {Process}.
\newblock \emph{Indiana University Mathematics Journal}, 6(4): 679--684.

\bibitem[{Benjamins et~al.(2022)Benjamins, Eimer, Schubert, Mohan, Biedenkapp, Rosenhahn, Hutter, and Lindauer}]{benjamins_contextualize_2022}
Benjamins, C.; Eimer, T.; Schubert, F.; Mohan, A.; Biedenkapp, A.; Rosenhahn, B.; Hutter, F.; and Lindauer, M. 2022.
\newblock Contextualize {Me} -- {The} {Case} for {Context} in {Reinforcement} {Learning}.
\newblock ArXiv:2202.04500 [cs].

\bibitem[{Brockman et~al.(2016)Brockman, Cheung, Pettersson, Schneider, Schulman, Tang, and Zaremba}]{brockman_openai_2016}
Brockman, G.; Cheung, V.; Pettersson, L.; Schneider, J.; Schulman, J.; Tang, J.; and Zaremba, W. 2016.
\newblock {OpenAI} {Gym}.
\newblock ArXiv:1606.01540 [cs].

\bibitem[{Camacho et~al.(2021)Camacho, Varley, Zeng, Jain, Iscen, and Kalashnikov}]{camacho_reward_2021}
Camacho, A.; Varley, J.; Zeng, A.; Jain, D.; Iscen, A.; and Kalashnikov, D. 2021.
\newblock Reward {Machines} for {Vision}-{Based} {Robotic} {Manipulation}.
\newblock In \emph{2021 {IEEE} {International} {Conference} on {Robotics} and {Automation} ({ICRA})}, 14284--14290. Xi'an, China: IEEE.
\newblock ISBN 978-1-72819-077-8.

\bibitem[{Chen, Xu, and Agrawal(2021)}]{chen_system_2021}
Chen, T.; Xu, J.; and Agrawal, P. 2021.
\newblock A {System} for {General} {In}-{Hand} {Object} {Re}-{Orientation}.
\newblock In \emph{Proceedings of the 5th {Annual} {Conference} on {Robot} {Learning}}.

\bibitem[{Danesh and Fern(2022)}]{danesh_out--distribution_2022}
Danesh, M.~H.; and Fern, A. 2022.
\newblock Out-of-{Distribution} {Dynamics} {Detection}: {RL}-{Relevant} {Benchmarks} and {Results}.
\newblock In \emph{Workshop on {Uncertainty} and {Robustness} in {Deep} {Learning} @ {ICML2021}}.
\newblock ArXiv:2107.04982 [cs].

\bibitem[{Danesh et~al.(2021)Danesh, Koul, Fern, and Khorram}]{danesh_re-understanding_2021}
Danesh, M.~H.; Koul, A.; Fern, A.; and Khorram, S. 2021.
\newblock Re-understanding {Finite}-{State} {Representations} of {Recurrent} {Policy} {Networks}.
\newblock In \emph{Proceedings of the 38th {International} {Conference} on {Machine} {Learning}}, 2388--2397. PMLR.
\newblock ISSN: 2640-3498.

\bibitem[{Dorfman, Shenfeld, and Tamar(2021)}]{dorfman_offline_2021}
Dorfman, R.; Shenfeld, I.; and Tamar, A. 2021.
\newblock Offline {Meta} {Reinforcement} {Learning} – {Identifiability} {Challenges} and {Effective} {Data} {Collection} {Strategies}.
\newblock In \emph{Advances in {Neural} {Information} {Processing} {Systems}}, volume~34, 4607--4618. Curran Associates, Inc.

\bibitem[{Duan et~al.(2016)Duan, Schulman, Chen, Bartlett, Sutskever, and Abbeel}]{duan_rl2_2016}
Duan, Y.; Schulman, J.; Chen, X.; Bartlett, P.~L.; Sutskever, I.; and Abbeel, P. 2016.
\newblock {RL}\${\textasciicircum}2\$: {Fast} {Reinforcement} {Learning} via {Slow} {Reinforcement} {Learning}.
\newblock \emph{arXiv:1611.02779 [cs, stat]}.
\newblock ArXiv: 1611.02779.

\bibitem[{Dunion et~al.(2023{\natexlab{a}})Dunion, McInroe, Luck, Hanna, and Albrecht}]{dunion_conditional_2023}
Dunion, M.; McInroe, T.; Luck, K.~S.; Hanna, J.; and Albrecht, S.~V. 2023{\natexlab{a}}.
\newblock Conditional {Mutual} {Information} for {Disentangled} {Representations} in {Reinforcement} {Learning}.
\newblock In \emph{Conference on {Neural} {Information} {Processing} {Systems}}.

\bibitem[{Dunion et~al.(2023{\natexlab{b}})Dunion, McInroe, Luck, Hanna, and Albrecht}]{dunion_temporal_2023}
Dunion, M.; McInroe, T.; Luck, K.~S.; Hanna, J.; and Albrecht, S.~V. 2023{\natexlab{b}}.
\newblock Temporal {Disentanglement} of {Representations} for {Improved} {Generalisation} in {Reinforcement} {Learning}.
\newblock In \emph{International {Conference} on {Learning} {Representations} ({ICLR})}.

\bibitem[{Finn, Abbeel, and Levine(2017)}]{finn_model-agnostic_2017}
Finn, C.; Abbeel, P.; and Levine, S. 2017.
\newblock Model-{Agnostic} {Meta}-{Learning} for {Fast} {Adaptation} of {Deep} {Networks}.
\newblock \emph{arXiv:1703.03400 [cs]}.
\newblock ArXiv: 1703.03400.

\bibitem[{Gupta et~al.(2022)Gupta, Pacchiano, Zhai, Kakade, and Levine}]{gupta_unpacking_2022}
Gupta, A.; Pacchiano, A.; Zhai, Y.; Kakade, S.~M.; and Levine, S. 2022.
\newblock Unpacking {Reward} {Shaping}: {Understanding} the {Benefits} of {Reward} {Engineering} on {Sample} {Complexity}.
\newblock ArXiv:2210.09579 [cs].

\bibitem[{Hallak, Di~Castro, and Mannor(2015)}]{hallak_contextual_2015}
Hallak, A.; Di~Castro, D.; and Mannor, S. 2015.
\newblock Contextual {Markov} {Decision} {Processes}.
\newblock \emph{arXiv:1502.02259 [cs, stat]}.
\newblock ArXiv: 1502.02259.

\bibitem[{Illanes et~al.(2020)Illanes, Yan, Icarte, and McIlraith}]{illanes_symbolic_2020}
Illanes, L.; Yan, X.; Icarte, R.~T.; and McIlraith, S.~A. 2020.
\newblock Symbolic {Plans} as {High}-{Level} {Instructions} for {Reinforcement} {Learning}.
\newblock \emph{Proceedings of the International Conference on Automated Planning and Scheduling}, 30: 540--550.

\bibitem[{Kingma and Ba(2015)}]{kingma_adam_2015}
Kingma, D.~P.; and Ba, J. 2015.
\newblock Adam: {A} {Method} for {Stochastic} {Optimization}.
\newblock In Bengio, Y.; and LeCun, Y., eds., \emph{3rd {International} {Conference} on {Learning} {Representations}, {ICLR} 2015, {San} {Diego}, {CA}, {USA}, {May} 7-9, 2015, {Conference} {Track} {Proceedings}}.

\bibitem[{Kirk et~al.(2021)Kirk, Zhang, Grefenstette, and Rocktäschel}]{kirk_survey_2021}
Kirk, R.; Zhang, A.; Grefenstette, E.; and Rocktäschel, T. 2021.
\newblock A {Survey} of {Generalisation} in {Deep} {Reinforcement} {Learning}.
\newblock \emph{arXiv:2111.09794 [cs]}.
\newblock ArXiv: 2111.09794.

\bibitem[{Langford(2017)}]{langford_contextual_2017}
Langford, J. 2017.
\newblock Contextual reinforcement learning.
\newblock In \emph{2017 {IEEE} {International} {Conference} on {Big} {Data} ({Big} {Data})}, 3--3.

\bibitem[{Langford(2018)}]{langford_real_2018}
Langford, J. 2018.
\newblock Real {World} {Reinforcement} {Learning}.

\bibitem[{Leike et~al.(2017)Leike, Martic, Krakovna, Ortega, Everitt, Lefrancq, Orseau, and Legg}]{leike_ai_2017}
Leike, J.; Martic, M.; Krakovna, V.; Ortega, P.~A.; Everitt, T.; Lefrancq, A.; Orseau, L.; and Legg, S. 2017.
\newblock {AI} {Safety} {Gridworlds}.
\newblock ArXiv:1711.09883 [cs].

\bibitem[{Mnih et~al.(2015)Mnih, Kavukcuoglu, Silver, Rusu, Veness, Bellemare, Graves, Riedmiller, Fidjeland, Ostrovski, Petersen, Beattie, Sadik, Antonoglou, King, Kumaran, Wierstra, Legg, and Hassabis}]{mnih_human-level_2015}
Mnih, V.; Kavukcuoglu, K.; Silver, D.; Rusu, A.~A.; Veness, J.; Bellemare, M.~G.; Graves, A.; Riedmiller, M.; Fidjeland, A.~K.; Ostrovski, G.; Petersen, S.; Beattie, C.; Sadik, A.; Antonoglou, I.; King, H.; Kumaran, D.; Wierstra, D.; Legg, S.; and Hassabis, D. 2015.
\newblock Human-level control through deep reinforcement learning.
\newblock \emph{Nature}, 518(7540): 529--533.
\newblock Number: 7540 Publisher: Nature Publishing Group.

\bibitem[{Ng, Harada, and Russell(1999)}]{ng_policy_1999}
Ng, A.~Y.; Harada, D.; and Russell, S. 1999.
\newblock Policy invariance under reward transformations: {Theory} and application to reward shaping.
\newblock In \emph{In {Proceedings} of the {Sixteenth} {International} {Conference} on {Machine} {Learning}}, 278--287. Morgan Kaufmann.

\bibitem[{Schrittwieser et~al.(2020)Schrittwieser, Antonoglou, Hubert, Simonyan, Sifre, Schmitt, Guez, Lockhart, Hassabis, Graepel, Lillicrap, and Silver}]{schrittwieser_mastering_2020}
Schrittwieser, J.; Antonoglou, I.; Hubert, T.; Simonyan, K.; Sifre, L.; Schmitt, S.; Guez, A.; Lockhart, E.; Hassabis, D.; Graepel, T.; Lillicrap, T.; and Silver, D. 2020.
\newblock Mastering {Atari}, {Go}, chess and shogi by planning with a learned model.
\newblock \emph{Nature}, 588(7839): 604--609.
\newblock Number: 7839 Publisher: Nature Publishing Group.

\bibitem[{Shrestha et~al.(2020)Shrestha, Lee, Tadepalli, and Fern}]{shrestha_deepaveragers_2020}
Shrestha, A.~K.; Lee, S.; Tadepalli, P.; and Fern, A. 2020.
\newblock {DeepAveragers}: {Offline} {Reinforcement} {Learning} {By} {Solving} {Derived} {Non}-{Parametric} {MDPs}.
\newblock In \emph{International {Conference} on {Learning} {Representations}}.

\bibitem[{Sutton, Precup, and Singh(1999)}]{sutton_between_1999-1}
Sutton, R.~S.; Precup, D.; and Singh, S. 1999.
\newblock Between {MDPs} and semi-{MDPs}: {A} framework for temporal abstraction in reinforcement learning.
\newblock \emph{Artificial intelligence}, 112(1-2): 181--211.
\newblock Publisher: Elsevier.

\bibitem[{Tamar et~al.(2016)Tamar, WU, Thomas, Levine, and Abbeel}]{tamar_value_2016}
Tamar, A.; WU, Y.; Thomas, G.; Levine, S.; and Abbeel, P. 2016.
\newblock Value {Iteration} {Networks}.
\newblock In \emph{Advances in {Neural} {Information} {Processing} {Systems}}, volume~29. Curran Associates, Inc.

\bibitem[{Taylor and Stone(2009)}]{taylor_transfer_2009}
Taylor, M.~E.; and Stone, P. 2009.
\newblock Transfer {Learning} for {Reinforcement} {Learning} {Domains}: {A} {Survey}.
\newblock \emph{Journal of Machine Learning Research}, 10(56): 1633--1685.

\bibitem[{Terry et~al.(2021)Terry, Black, Grammel, Jayakumar, Hari, Sullivan, Santos, Perez, Horsch, Dieffendahl, Williams, Lokesh, and Ravi}]{terry_pettingzoo_2021}
Terry, J.~K.; Black, B.; Grammel, N.; Jayakumar, M.; Hari, A.; Sullivan, R.; Santos, L.; Perez, R.; Horsch, C.; Dieffendahl, C.; Williams, N.~L.; Lokesh, Y.; and Ravi, P. 2021.
\newblock {PettingZoo}: {Gym} for {Multi}-{Agent} {Reinforcement} {Learning}.
\newblock ArXiv:2009.14471 [cs, stat].

\bibitem[{Toro~Icarte et~al.(2018)Toro~Icarte, Klassen, Valenzano, and McIlraith}]{toro_icarte_using_2018}
Toro~Icarte, R.; Klassen, T.; Valenzano, R.; and McIlraith, S. 2018.
\newblock Using {Reward} {Machines} for {High}-{Level} {Task} {Specification} and {Decomposition} in {Reinforcement} {Learning}.
\newblock In \emph{Proceedings of the 35th {International} {Conference} on {Machine} {Learning}}, 2107--2116. PMLR.
\newblock ISSN: 2640-3498.

\bibitem[{Toro~Icarte et~al.(2022)Toro~Icarte, Klassen, Valenzano, and McIlraith}]{toro_icarte_reward_2022-1}
Toro~Icarte, R.; Klassen, T.~Q.; Valenzano, R.; and McIlraith, S.~A. 2022.
\newblock Reward {Machines}: {Exploiting} {Reward} {Function} {Structure} in {Reinforcement} {Learning}.
\newblock \emph{Journal of Artificial Intelligence Research}, 73: 173--208.

\bibitem[{Toro~Icarte et~al.(2019)Toro~Icarte, Waldie, Klassen, Valenzano, Castro, and McIlraith}]{toro_icarte_learning_2019}
Toro~Icarte, R.; Waldie, E.; Klassen, T.; Valenzano, R.; Castro, M.; and McIlraith, S. 2019.
\newblock Learning {Reward} {Machines} for {Partially} {Observable} {Reinforcement} {Learning}.
\newblock In \emph{Advances in {Neural} {Information} {Processing} {Systems}}, volume~32. Curran Associates, Inc.

\bibitem[{Torrey and Shavlik(2009)}]{torrey_chapter_2009}
Torrey, L.~A.; and Shavlik, J. 2009.
\newblock Chapter 11 {Transfer} {Learning}.
\newblock In \emph{Handbook of {Research} on {Machine} {Learning} {Applications} and {Trends}: {Algorithms}, {Methods}, and {Techniques}}, 242--264. IGI Global.

\bibitem[{Tseng et~al.(2017)Tseng, Luo, Cui, Chien, Ten~Haken, and Naqa}]{tseng_deep_2017}
Tseng, H.-H.; Luo, Y.; Cui, S.; Chien, J.-T.; Ten~Haken, R.~K.; and Naqa, I.~E. 2017.
\newblock Deep reinforcement learning for automated radiation adaptation in lung cancer.
\newblock \emph{Medical Physics}, 44(12): 6690--6705.

\bibitem[{Vaezipoor et~al.(2021)Vaezipoor, Li, Icarte, and Mcilraith}]{vaezipoor_ltl2action_2021}
Vaezipoor, P.; Li, A.~C.; Icarte, R. A.~T.; and Mcilraith, S.~A. 2021.
\newblock {LTL2Action}: {Generalizing} {LTL} {Instructions} for {Multi}-{Task} {RL}.
\newblock In \emph{Proceedings of the 38th {International} {Conference} on {Machine} {Learning}}, 10497--10508. PMLR.
\newblock ISSN: 2640-3498.

\bibitem[{Wang et~al.(2017)Wang, Kurth-Nelson, Tirumala, Soyer, Leibo, Munos, Blundell, Kumaran, and Botvinick}]{wang_learning_2017}
Wang, J.~X.; Kurth-Nelson, Z.; Tirumala, D.; Soyer, H.; Leibo, J.~Z.; Munos, R.; Blundell, C.; Kumaran, D.; and Botvinick, M. 2017.
\newblock Learning to reinforcement learn.
\newblock \emph{arXiv:1611.05763 [cs, stat]}.
\newblock ArXiv: 1611.05763.

\bibitem[{Zhang et~al.(2018)Zhang, Vinyals, Munos, and Bengio}]{zhang_study_2018}
Zhang, C.; Vinyals, O.; Munos, R.; and Bengio, S. 2018.
\newblock A {Study} on {Overfitting} in {Deep} {Reinforcement} {Learning}.
\newblock ArXiv:1804.06893 [cs, stat].

\bibitem[{Zintgraf et~al.(2019)Zintgraf, Shiarlis, Igl, Schulze, Gal, Hofmann, and Whiteson}]{zintgraf_varibad_2019}
Zintgraf, L.; Shiarlis, K.; Igl, M.; Schulze, S.; Gal, Y.; Hofmann, K.; and Whiteson, S. 2019.
\newblock {VariBAD}: {A} {Very} {Good} {Method} for {Bayes}-{Adaptive} {Deep} {RL} via {Meta}-{Learning}.
\newblock In \emph{International {Conference} on {Learning} {Representations}}.

\end{thebibliography}

\appendix
\section{PCG Experiments}\label{app:pcg}

\begin{figure*}[t]
    \centering
    \includegraphics[width=1\textwidth]{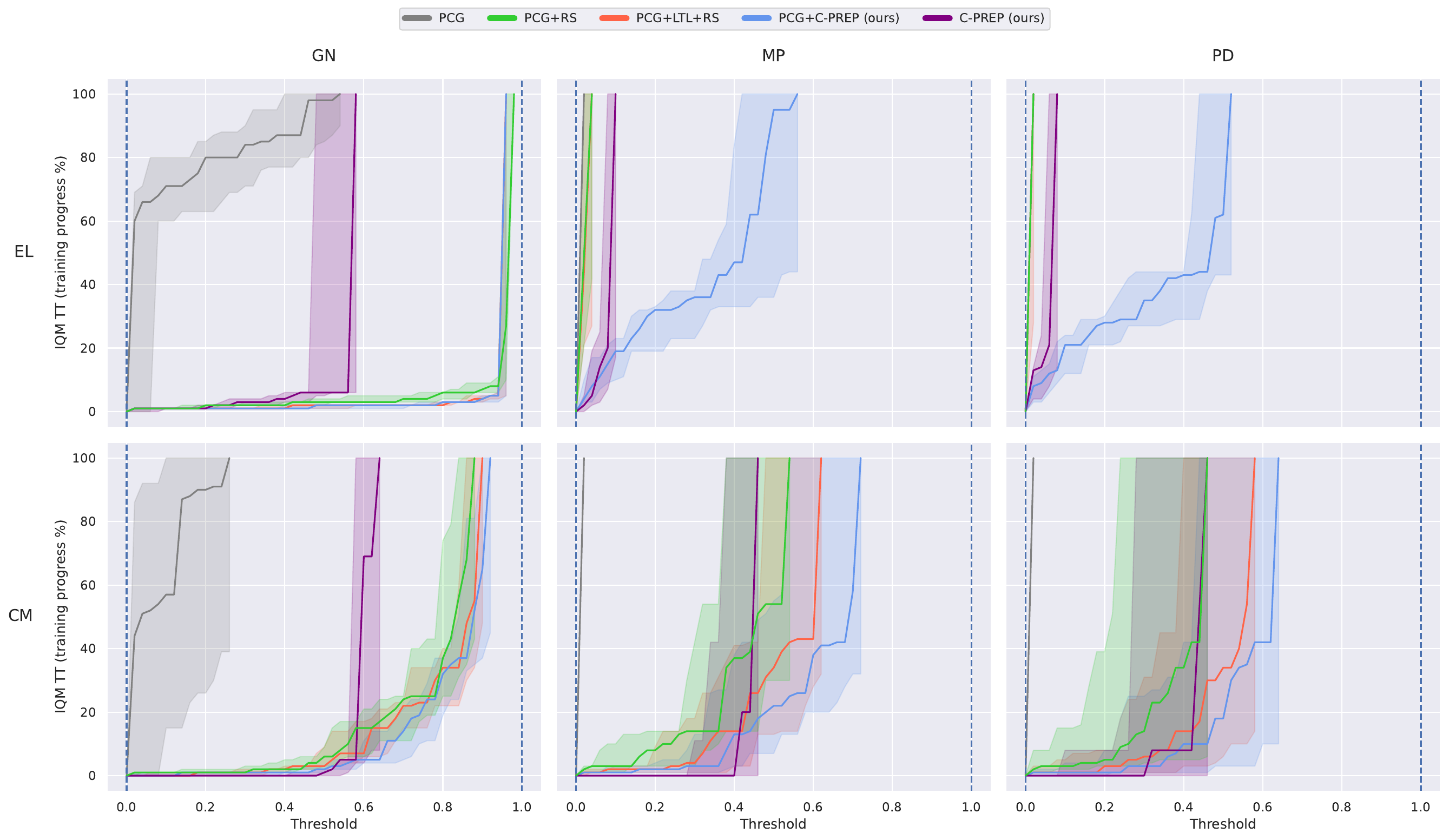}
    \caption{The IQM \ttt as a function of the threshold for configurations with \ctxPCG base context representations.}
    \label{fig:res:tsf-pcg}
\end{figure*}

\cref{tab:pcg} shows that among configurations that use \ctxPCG-based context representation, our \ctxCTL{}+\cPrep performed best in all \CMDP{}s of the GN, MP, and PD environments in the $\ttt_{\auc}$ utility. The near-zero \js values and small ($<0.1$) to negative \tr values indicate overfitting and nonbeneficial transfer in all \CMDP{}s for all configurations using \ctxPCG. \cPrep without \ctxPCG shows positive (or negative but close to zero) \tr in all configurations. \cref{fig:res:tsf-pcg} presents the IQM \ttt as a function of the threshold for comparison with the results using \ctxCTL representations shown in \cref{fig:multi-ttt}.

\begin{table*}[t]
    \centering
    \resizebox{1\textwidth}{!}{
    \bgroup
    \def\arraystretch{1.2}
    \begin{tabular}{|c|c|c?c|c|c|c|c|c|c|}
\hline
 Utility   & Context Space   & Environment   & PCG                  & PCG+RS             & PCG+LTL+RS         & PCG+C-PREP (ours)    & C-PREP (ours)       \\
\hline
 \Xhline{4\arrayrulewidth}
\parbox[t]{2mm}{\multirow{7}{*}{\rotatebox[origin=c]{90}{$\ttt_{\auc}$}}}       &  \multirow{3}{*}{EL}  & GN   & 84.95 +- 3.23                  & 7.12 +- 0.66                   & 5.89 +- 0.22                   & {\bf 5.82 +- 0.28}             & 42.15 +- 3.95                        \\
        &    &  MP   & 98.04 +- 0.00                  & 96.11 +- 1.14                  & 95.02 +- 2.03                  & {\bf 61.96 +- 9.37}            & 86.43 +- 1.89                        \\
        &    & PD   & 98.04 +- 0.00                  & 98.04 +- 0.00                  & 96.75 +- 0.90                  & {\bf 58.97 +- 6.69}            & 88.35 +- 1.59                        \\
        \cline{2-8}
        &  \multirow{3}{*}{CM}    & GN   & 85.61 +- 12.21                 & 20.01 +- 3.63                  & 16.45 +- 2.37                  & {\bf 14.58 +- 2.19}            & 32.98 +- 0.68                        \\
        &      &   MP  & 98.04 +- 0.00                  & 51.29 +- 12.97                 & 42.41 +- 12.14                 & {\bf 32.73 +- 13.21}           & 47.74 +- 11.46                       \\
        &      & PD   & 98.04 +- 0.00                  & 55.94 +- 19.14                 & 43.39 +- 16.48                 & {\bf 39.95 +- 18.23}           & 51.02 +- 15.46                       \\
        \cline{2-8}
        & PO    & ON   & 98.04 +- 0.00                  & 92.56 +- 13.58                 & 93.01 +- 3.80                  & 37.65 +- 22.90                 & {\bf 22.54 +- 1.57}                  \\
 \Xhline{4\arrayrulewidth}
\parbox[t]{2mm}{\multirow{7}{*}{\rotatebox[origin=c]{90}{\js}}}        &  \multirow{3}{*}{EL}  & GN   & {\bf 0.01 +- 0.03}             & 0.00 +- 0.01                   & 0.00 +- 0.01                   & 0.00 +- 0.01                   & 0.01 +- 0.05                         \\
         &    &  MP   & 0.00 +- 0.00                   & 0.00 +- 0.00                   & 0.00 +- 0.00                   & 0.00 +- 0.00                   & {\bf 0.01 +- 0.02}                   \\
         &    & PD   & {\bf 0.00 +- 0.00}             & {\bf 0.00 +- 0.00}             & {\bf 0.00 +- 0.00}             & {\bf 0.00 +- 0.00}             & 0.00 +- 0.01                         \\
         \cline{2-8}
         &  \multirow{3}{*}{CM}    & GN   & 0.00 +- 0.05                   & 0.00 +- 0.02                   & 0.18 +- 0.14                   & 0.10 +- 0.18                   & {\bf 0.49 +- 0.06}                   \\
         &      &  MP   & 0.00 +- 0.00                   & 0.00 +- 0.00                   & 0.00 +- 0.02                   & 0.00 +- 0.00                   & {\bf 0.42 +- 0.10}                   \\
         &      & PD   & 0.00 +- 0.00                   & 0.00 +- 0.00                   & 0.00 +- 0.00                   & 0.00 +- 0.13                   & {\bf 0.31 +- 0.19}                   \\
         \cline{2-8}
         & PO    & ON   & 0.00 +- 0.00                   & 0.00 +- 0.00                   & 0.00 +- 0.00                   & 0.00 +- 0.00                   & {\bf 0.75 +- 0.02}                   \\
 \Xhline{4\arrayrulewidth}
\parbox[t]{2mm}{\multirow{7}{*}{\rotatebox[origin=c]{90}{\tr}}}        &  \multirow{3}{*}{EL}  & GN   & \textit{-0.86 +- 0.05} & 0.05 +- 0.01                   & {\bf 0.05 +- 0.02}             & 0.04 +- 0.02                   & 0.04 +- 0.02                         \\
         &    &  MP   & \textit{-1.00 +- 0.00} & \textit{-0.85 +- 0.03} & \textit{-0.93 +- 0.13} & \textit{-0.31 +- 0.21} & {\bf 0.06 +- 0.11}                   \\
         &    & PD   & \textit{-1.00 +- 0.00} & \textit{-0.84 +- 0.09} & \textit{-0.94 +- 0.07} & \textit{-0.33 +- 0.16} & \textit{{\bf -0.05 +- 0.12}} \\
         \cline{2-8}
         &  \multirow{3}{*}{CM}    & GN   & \textit{-0.77 +- 0.37} & \textit{-0.02 +- 0.06} & \textit{-0.02 +- 0.04} & 0.00 +- 0.02                   & {\bf 0.08 +- 0.04}                   \\
         &      &  MP   & \textit{-1.00 +- 0.00} & {\bf 0.11 +- 0.18}             & \textit{-0.03 +- 0.08} & \textit{-0.02 +- 0.03} & 0.07 +- 0.05                         \\
         &      & PD   & \textit{-0.60 +- 0.49} & {\bf 0.21 +- 0.21}             & 0.06 +- 0.19                   & 0.06 +- 0.06                   & 0.07 +- 0.07                         \\
         \cline{2-8}
         & PO    & ON   & 0.00 +- 0.00                   & inf\tablefootnote{Target policy scored 0 on all seeds.}                     & \textit{-0.71 +- 1.79} & 0.31 +- 0.55                   & {\bf 0.31 +- 0.03}                   \\
\hline
\end{tabular}
\egroup
}
\caption{IQM and standard deviation transfer utilities in tested \CMDP{}S using \ctxPCG as the base context representation. The best results for each CMDP (row) are marked in bold. Negative \tr values that indicate non-beneficial transfer are italicized.}
    \label{tab:pcg}
\end{table*}

\section{Ablation Tests}\label{app:abl}
\begin{figure*}[t]
    \centering
    \begin{minipage}{0.65\textwidth}
        \includegraphics[width=1\textwidth]{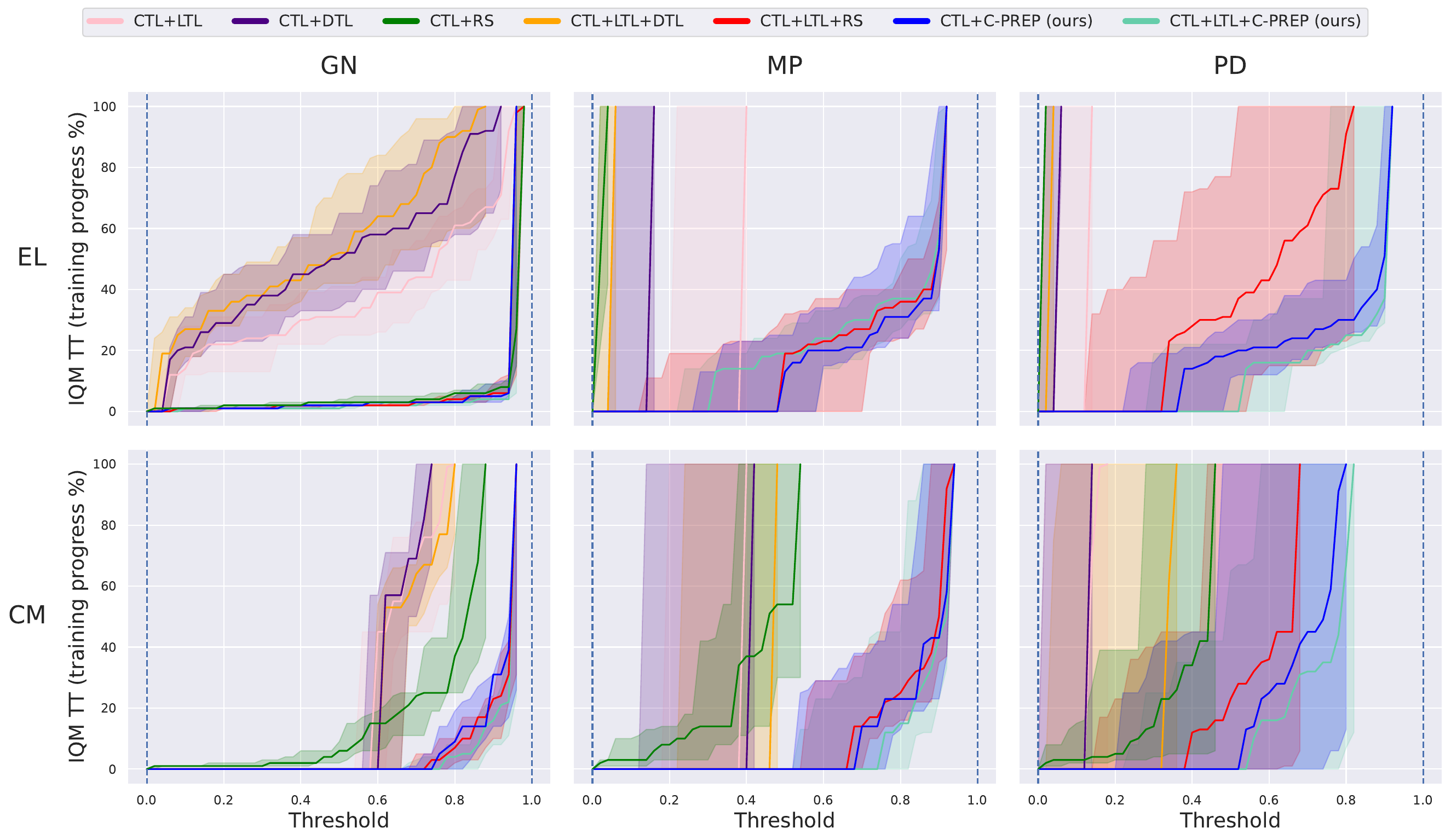}
        \subcaption{Ablation configurations using \ctxCTL as the base context representation}
        \label{fig:multi-ttt-abl:abl}
    \end{minipage}\hfill
    \begin{minipage}{0.33\textwidth}
        \includegraphics[width=1\textwidth]{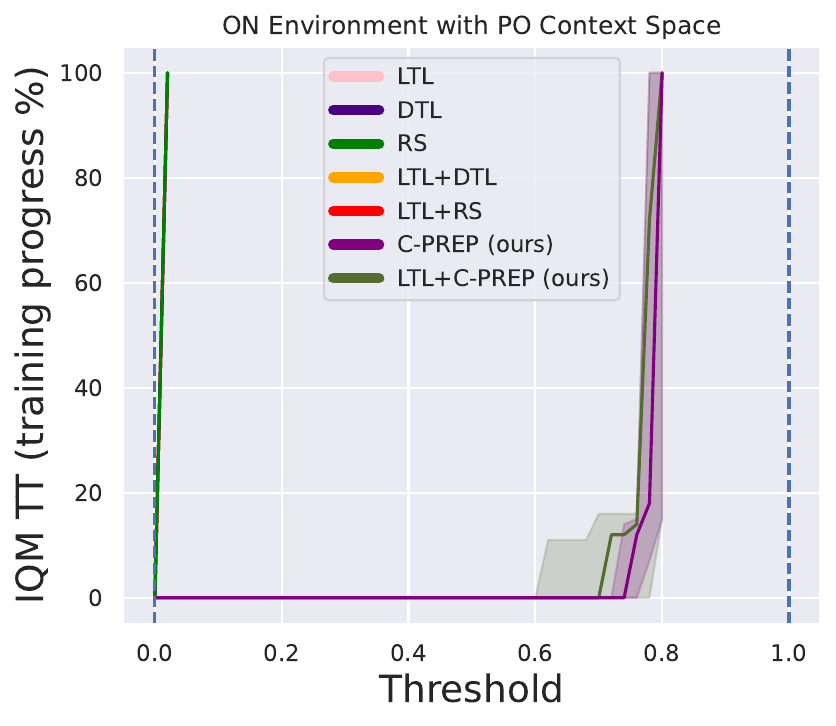}
        \subcaption{Ablation configurations using \ctxCTL as the base context representation}
        \label{fig:multi-ttt-abl:on-abl}
    \end{minipage}
    
    \caption{The IQM \ttt as a function of the threshold for ablation test configurations.}
    \label{fig:multi-ttt-abl}
\end{figure*}

\cref{tab:abl} shows the results of our ablation tests based on the \cPrep components. In the GN, MP, and PD environments, we used \ctxCTL as the base context representation. The results show that all components are necessary to achieve the highest transfer performance. As can be seen in \cref{fig:multi-ttt-abl:abl}, configurations that do not use \modRS achieve low performance thresholds in most tasks. We further observe that adding \modLTL to our representation (\ctxCTL{}+\modLTL{}+\cPrep) outperforms \ctxCTL{}+\cPrep in the $\ttt_{\auc}$ utility in five out of six of the \CMDP{}s in these environments. In the ON environment, due to the success of \cPrep without \ctxCTL, we additionally tested ablations in this environment without \ctxCTL. From the ablation analysis in \cref{fig:multi-ttt-abl:on-abl} we see that without all \cPrep components, the agent cannot complete the tasks after transfer, which shows the significance of each component. In this setting (ON) \modLTL{}+\cPrep performs on par with \cPrep, showing that our representation is sufficient.

\begin{sidewaystable*}

    \centering
    \resizebox{1\textwidth}{!}{
    \bgroup
    \def\arraystretch{1.2}
    \begin{tabular}{|c|c|c?c|c|c|c|c|c|c??c|c|c|c|c|c|c|}
        \hline
         Utility   & Context Space   & Environment   & CTL+LTL                        & CTL+DTL                        & CTL+RS                         & CTL+LTL+DTL                    & CTL+LTL+RS                     & CTL+C-PREP (ours)    & CTL+LTL+C-PREP (ours)          & PCG+LTL                        & PCG+DTL                        & PCG+RS                         & PCG+LTL+DTL                                           & PCG+LTL+RS                     & PCG+C-PREP (ours)                    & PCG+LTL+C-PREP (ours)         \\
        \Xhline{4\arrayrulewidth}
\parbox[t]{2mm}{\multirow{7}{*}{\rotatebox[origin=c]{90}{$\ttt_{\auc}$}}}       &  \multirow{3}{*}{EL}  & GN   & 35.25 +- 6.97                  & 48.41 +- 10.33                 & 7.12 +- 0.66                   & 53.00 +- 10.25                 & 6.21 +- 0.42                   & 6.42 +- 0.44         & {\bf 5.69 +- 0.24}             & 80.30 +- 8.29                  & 72.21 +- 15.60                 & 7.12 +- 0.66                   & {\bf 5.56 +- 0.15}                   & 5.89 +- 0.22                   & 5.82 +- 0.28                         & 84.37 +- 9.77                 \\
            &   & MP   & 62.09 +- 15.17                 & 84.31 +- 8.59                  & 96.11 +- 1.14                  & 94.12 +- 12.98                 & 19.88 +- 9.36                  & {\bf 18.32 +- 9.27}  & 24.93 +- 6.72                  & 98.04 +- 0.00                  & 98.04 +- 0.00                  & 96.11 +- 1.14                  & {\bf 54.92 +- 9.86}                  & 95.02 +- 2.03                  & 61.96 +- 9.37                        & 98.04 +- 0.00                 \\
            &   & PD   & 85.62 +- 6.37                  & 95.42 +- 2.08                  & 98.04 +- 0.00                  & 96.08 +- 1.75                  & 37.58 +- 23.38                 & 21.68 +- 6.55        & {\bf 17.07 +- 15.58}           & 98.04 +- 0.00                  & 98.04 +- 0.00                  & 98.04 +- 0.00                  & {\bf 43.21 +- 12.76}                 & 96.75 +- 0.90                  & 58.97 +- 6.69                        & 98.04 +- 0.00                 \\
            \cline{2-17}
            & \multirow{3}{*}{CM}    & GN   & 30.94 +- 4.47                  & 31.11 +- 2.80                  & 20.01 +- 3.63                  & 29.73 +- 3.06                  & 7.14 +- 1.41                   & 7.64 +- 1.72         & {\bf 6.46 +- 1.33}             & 80.90 +- 5.40                  & 91.53 +- 3.80                  & 20.01 +- 3.63                  & {\bf 13.75 +- 2.15}                  & 16.45 +- 2.37                  & 14.58 +- 2.19                        & 85.39 +- 7.17                 \\
            &     & MP   & 59.05 +- 13.28                 & 53.03 +- 19.86                 & 51.29 +- 12.97                 & 52.37 +- 20.12                 & 14.24 +- 8.85                  & 13.64 +- 8.60        & {\bf 11.58 +- 11.46}           & 93.42 +- 4.32                  & 98.04 +- 0.00                  & 51.29 +- 12.97                 & 33.52 +- 12.62                       & 42.41 +- 12.14                 & {\bf 32.73 +- 13.21}                 & 97.11 +- 3.67                 \\
            &     & PD   & 82.42 +- 21.06                 & 86.27 +- 15.43                 & 55.94 +- 19.14                 & 64.64 +- 24.99                 & 37.52 +- 25.46                 & 28.95 +- 24.15       & {\bf 25.22 +- 20.73}           & 98.04 +- 2.83                  & 98.04 +- 0.00                  & 55.94 +- 19.14                 & 40.65 +- 18.64                       & 43.39 +- 16.48                 & {\bf 39.95 +- 18.23}                 & 98.01 +- 2.92                 \\
            \cline{2-17}
            & PO    & ON   & 98.04 +- 0.00                  & 98.04 +- 0.00                  & 92.56 +- 13.58                 & 98.04 +- 0.00                  & 97.50 +- 5.59                  & {\bf 32.93 +- 10.21} & 52.05 +- 33.84                 & 98.04 +- 0.00                  & 98.04 +- 0.00                  & 92.56 +- 13.58                 & {\bf 34.35 +- 6.34}                  & 93.01 +- 3.80                  & 37.65 +- 22.90                       & 98.04 +- 0.00                 \\
         \Xhline{4\arrayrulewidth}
\parbox[t]{2mm}{\multirow{7}{*}{\rotatebox[origin=c]{90}{\js}}}       &  \multirow{3}{*}{EL}  & GN   & 0.05 +- 0.03                   & 0.06 +- 0.02                   & 0.00 +- 0.01                   & 0.04 +- 0.03                   & 0.07 +- 0.09                   & 0.05 +- 0.05         & {\bf 0.10 +- 0.06}             & 0.01 +- 0.01                   & 0.01 +- 0.02                   & 0.00 +- 0.01                   & {\bf 0.03 +- 0.03}                   & 0.00 +- 0.01                   & 0.00 +- 0.01                         & 0.02 +- 0.02                  \\
                 &   & MP   & 0.38 +- 0.15                   & 0.15 +- 0.08                   & 0.00 +- 0.00                   & 0.04 +- 0.13                   & 0.48 +- 0.26                   & {\bf 0.49 +- 0.17}   & 0.31 +- 0.19                   & {\bf 0.00 +- 0.00}             & {\bf 0.00 +- 0.00}             & {\bf 0.00 +- 0.00}             & 0.00 +- 0.01                         & {\bf 0.00 +- 0.00}             & {\bf 0.00 +- 0.00}                   & {\bf 0.00 +- 0.00}            \\
                 &   & PD   & 0.13 +- 0.07                   & 0.04 +- 0.02                   & 0.00 +- 0.00                   & 0.04 +- 0.02                   & 0.34 +- 0.20                   & 0.38 +- 0.12         & {\bf 0.53 +- 0.19}             & {\bf 0.00 +- 0.00}             & {\bf 0.00 +- 0.00}             & {\bf 0.00 +- 0.00}             & {\bf 0.00 +- 0.00}                   & {\bf 0.00 +- 0.00}             & {\bf 0.00 +- 0.00}                   & {\bf 0.00 +- 0.00}            \\
                 \cline{2-17}
                 & \multirow{3}{*}{CM}    & GN   & 0.59 +- 0.05                   & 0.62 +- 0.05                   & 0.00 +- 0.02                   & 0.61 +- 0.04                   & 0.73 +- 0.05                   & 0.75 +- 0.08         & {\bf 0.76 +- 0.09}             & 0.00 +- 0.00                   & 0.00 +- 0.00                   & 0.00 +- 0.02                   & {\bf 0.25 +- 0.14}                   & 0.18 +- 0.14                   & 0.10 +- 0.18                         & 0.00 +- 0.06                  \\
                 &     & MP   & 0.39 +- 0.13                   & 0.41 +- 0.20                   & 0.00 +- 0.00                   & 0.46 +- 0.20                   & 0.66 +- 0.11                   & 0.68 +- 0.10         & {\bf 0.74 +- 0.17}             & {\bf 0.00 +- 0.00}             & {\bf 0.00 +- 0.00}             & {\bf 0.00 +- 0.00}             & 0.00 +- 0.06                         & 0.00 +- 0.02                   & {\bf 0.00 +- 0.00}                   & {\bf 0.00 +- 0.00}            \\
                 &     & PD   & 0.13 +- 0.19                   & 0.13 +- 0.14                   & 0.00 +- 0.00                   & 0.32 +- 0.23                   & 0.38 +- 0.27                   & 0.52 +- 0.25         & {\bf 0.55 +- 0.25}             & {\bf 0.00 +- 0.00}             & {\bf 0.00 +- 0.00}             & {\bf 0.00 +- 0.00}             & 0.00 +- 0.12                         & {\bf 0.00 +- 0.00}             & 0.00 +- 0.13                         & {\bf 0.00 +- 0.00}            \\
                 \cline{2-17}
                 & PO    & ON   & 0.00 +- 0.00                   & 0.00 +- 0.00                   & 0.00 +- 0.00                   & 0.00 +- 0.00                   & 0.00 +- 0.00                   & {\bf 0.14 +- 0.26}   & 0.01 +- 0.20                   & {\bf 0.00 +- 0.00}             & {\bf 0.00 +- 0.00}             & {\bf 0.00 +- 0.00}             & {\bf 0.00 +- 0.00}                   & {\bf 0.00 +- 0.00}             & {\bf 0.00 +- 0.00}                   & {\bf 0.00 +- 0.00}            \\
         \Xhline{4\arrayrulewidth}
\parbox[t]{2mm}{\multirow{7}{*}{\rotatebox[origin=c]{90}{\tr}}}       &  \multirow{3}{*}{EL}  & GN   & \textit{-0.24 +- 0.09} & \textit{-0.44 +- 0.14} & 0.05 +- 0.01                   & \textit{-0.48 +- 0.13} & {\bf 0.08 +- 0.03}             & 0.07 +- 0.03         & 0.06 +- 0.02                   & \textit{-0.80 +- 0.11} & \textit{-0.69 +- 0.20} & 0.05 +- 0.01                   & 0.05 +- 0.03                         & {\bf 0.05 +- 0.02}             & 0.04 +- 0.02                         & \textit{-0.86 +- 0.12}\\
         tr        &   & MP   & \textit{-0.96 +- 0.06} & \textit{-0.95 +- 0.04} & \textit{-0.85 +- 0.03} & \textit{-1.00 +- 0.01} & {\bf 0.24 +- 0.18}             & 0.16 +- 0.12         & 0.14 +- 0.07                   & \textit{-1.00 +- 0.00} & \textit{-1.00 +- 0.00} & \textit{-0.85 +- 0.03} & \textit{-0.32 +- 0.20}       & \textit{-0.93 +- 0.13} & \textit{{\bf -0.31 +- 0.21}} & \textit{-1.00 +- 0.00}\\
                 &   & PD   & \textit{-0.98 +- 0.01} & \textit{-0.99 +- 0.03} & \textit{-0.84 +- 0.09} & \textit{-1.00 +- 0.00} & \textit{-0.14 +- 0.38} & 0.14 +- 0.09         & {\bf 0.20 +- 0.25}             & \textit{-1.00 +- 0.00} & \textit{-1.00 +- 0.00} & \textit{-0.84 +- 0.09} & \textit{{\bf -0.10 +- 0.35}} & \textit{-0.94 +- 0.07} & \textit{-0.33 +- 0.16}       & \textit{-1.00 +- 0.00}\\
         \cline{2-17}
                 & \multirow{3}{*}{CM}    & GN   & \textit{-0.07 +- 0.07} & \textit{-0.09 +- 0.05} & \textit{-0.02 +- 0.06} & \textit{-0.07 +- 0.05} & {\bf 0.11 +- 0.01}             & 0.11 +- 0.01         & 0.10 +- 0.02                   & \textit{-0.58 +- 0.13} & \textit{-0.82 +- 0.09} & \textit{-0.02 +- 0.06} & \textit{-0.03 +- 0.04}       & \textit{-0.02 +- 0.04} & {\bf 0.00 +- 0.02}                   & \textit{-0.69 +- 0.14}\\
                 &     & MP   & \textit{-0.68 +- 0.28} & \textit{-0.71 +- 0.38} & 0.11 +- 0.18                   & \textit{-0.55 +- 0.34} & {\bf 0.26 +- 0.06}             & 0.21 +- 0.06         & 0.15 +- 0.05                   & \textit{-0.78 +- 0.30} & \textit{-1.00 +- 0.00} & {\bf 0.11 +- 0.18}             & \textit{-0.11 +- 0.04}       & \textit{-0.03 +- 0.08} & \textit{-0.02 +- 0.03}       & \textit{-0.92 +- 0.23}\\
                 &     & PD   & \textit{-0.77 +- 0.33} & \textit{-0.99 +- 0.19} & {\bf 0.21 +- 0.21}             & \textit{-0.46 +- 0.42} & \textit{-0.05 +- 0.20} & 0.06 +- 0.19         & 0.03 +- 0.22                   & \textit{-1.00 +- 0.39} & \textit{-0.99 +- 0.40} & {\bf 0.21 +- 0.21}             & \textit{-0.10 +- 0.07}       & 0.06 +- 0.19                   & 0.06 +- 0.06                         & \textit{-0.94 +- 0.48}\\
                 \cline{2-17}
                 & PO    & ON   & 0.00 +- 0.00                   & 0.00 +- 0.00                   & inf                     & 0.00 +- 0.00                   & \textit{-0.87 +- 0.90} & {\bf 0.69 +- 0.20}   & \textit{-0.08 +- 0.73} & 0.00 +- 0.00                   & 0.00 +- 0.00                   & inf                     & {\bf 0.69 +- 1.19}                   & \textit{-0.71 +- 1.79} & 0.31 +- 0.55                         & 0.00 +- 0.00                  \\
        \hline
        \end{tabular}
\egroup
}
\caption{IQM and standard deviation transfer utilities  for tested ablation configurations (environment-context space pairs) using both base context representation (\ctxCTL and \ctxPCG), aggregated over all seeds. The best results for each CMDP (row) and base context representation (left-right split) are marked in bold. Negative \tr values that indicate non-beneficial transfer are italicized.}
    \label{tab:abl}
\end{sidewaystable*}

\section{Context Representations}\label{app:ctx-reps}
\begin{figure*}[t]
    \centering
    \begin{minipage}[c][][t]{0.55\textwidth}
        \includegraphics[width=1\textwidth]{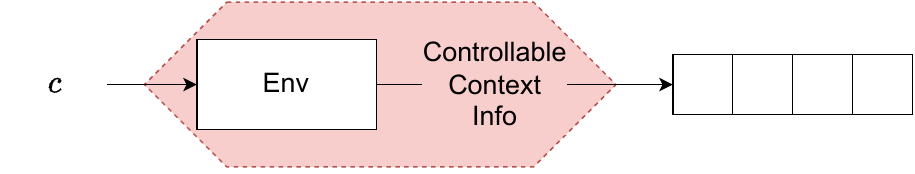}
    \end{minipage}\hfill
    \begin{minipage}[c][][t]{0.41\textwidth}
        \includegraphics[width=1\textwidth]{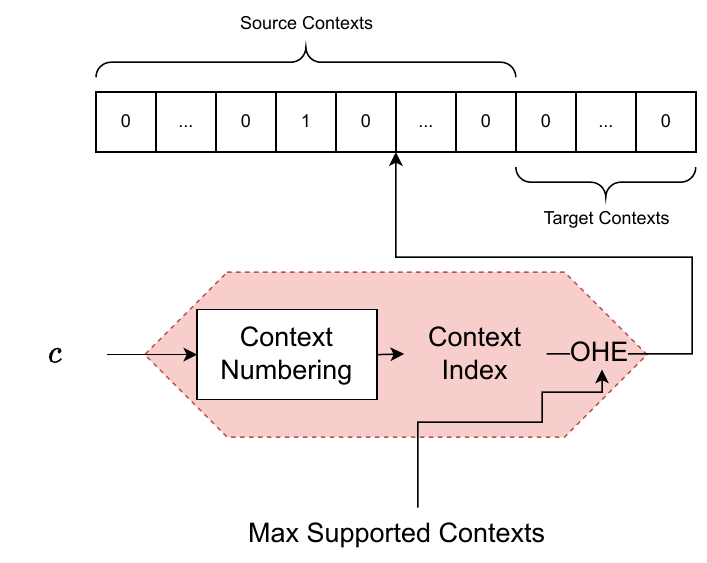}
    \end{minipage}
    \begin{minipage}[t][][t]{0.55\textwidth}
        \subcaption{}
        \label{fig:baselines:ctl}
    \end{minipage}\hfill
    \begin{minipage}[t][][t]{0.41\textwidth}
        \subcaption{}
        \label{fig:baselines:pcg}
    \end{minipage}
    \caption{Baseline context representation functions. (\subref{fig:baselines:ctl}) Controllable (\ctxCTL) environment context representation function. Each context is represented using the original informative context representation offered by the environment. (\subref{fig:baselines:pcg}) Procedural content generation (\ctxPCG) context representation function. Contexts have unique indices that are converted into binary indicator vectors (one-hot-encoding) for the source and target contexts together. One-hot encoding context representations are limited to a maximum number of supported contexts, determined ahead of time.}
    \label{fig:baselines}
\end{figure*}

In our experiments, we use both \ctxCTL and \ctxPCG context representations. \ctxCTL representations contain all the information required to generate the MDP induced by the context. In practice, this is part of the observation that is constant throughout the task. \ctxPCG representations are random indices assigned to each context as an identifier. These indices are provided to the agent as one-hot-encoding vectors. \cref{fig:baselines} illustrates the implementation of the corresponding context representation functions.

\begin{figure*}
    \centering
    \begin{minipage}[t][][t]{0.36\textwidth}
        \centering
        \includegraphics[width=0.48\textwidth]{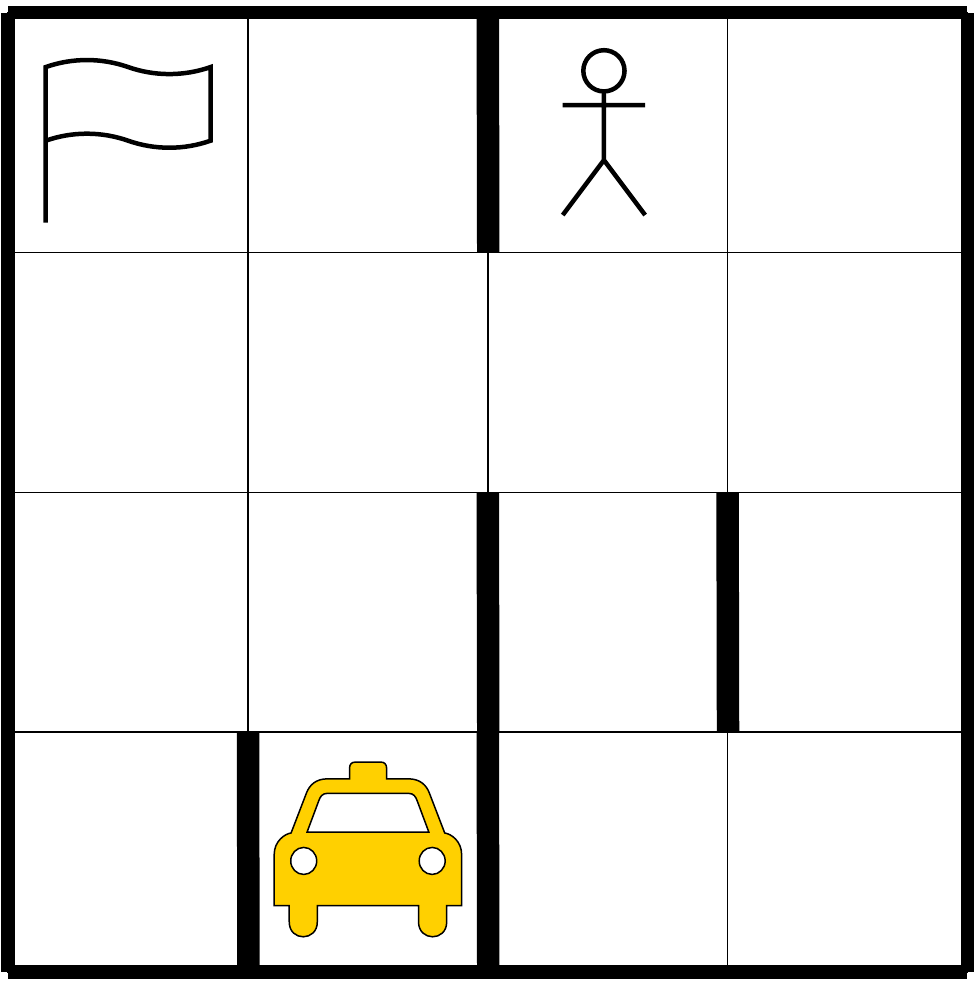}
        \hfill
        \includegraphics[width=0.48\textwidth]{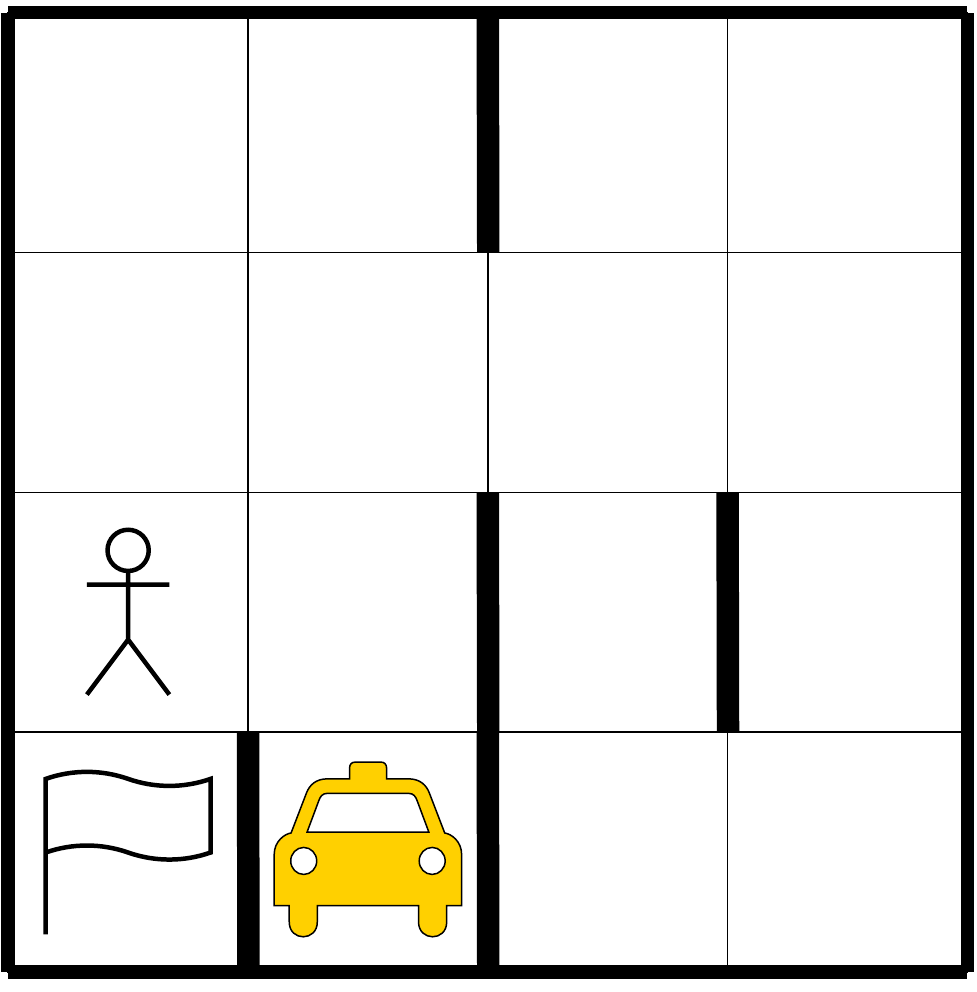}
        \subcaption{Two different EL contexts in the PD environment. The taxi icon is the acting agent, the person icon is the passenger to be picked up, and the flag icon is the passenger's destination. The agent can navigate to any adjacent cell, but cannot cross thick walls.}
        \label{fig:contexts}
    \end{minipage}\hfill
    \begin{minipage}[t][][t]{0.62\textwidth}
        \centering
        \includegraphics[width=0.28\textwidth]{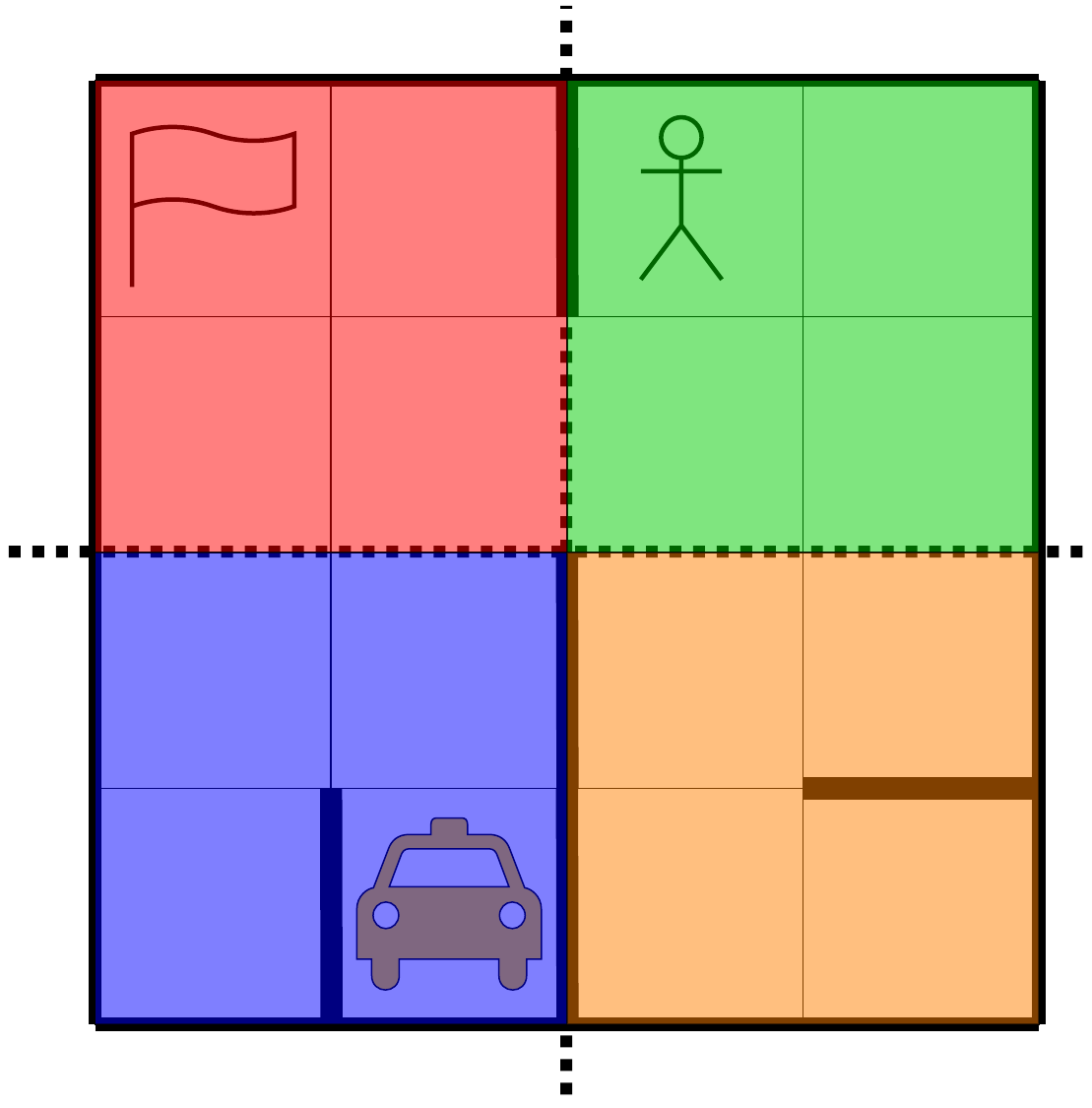}
        \includegraphics[width=0.7\textwidth]{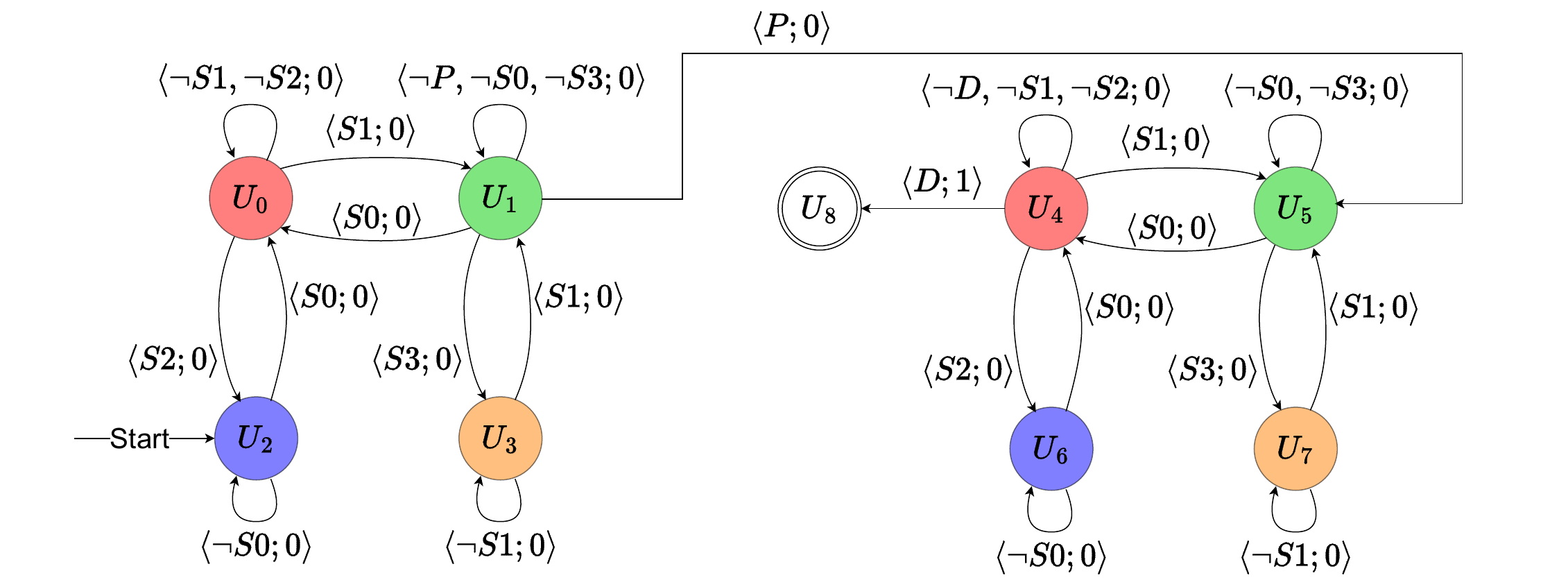}
        \cprotect\subcaption{An example abstraction of a task in the PD environment on a $4\times4$ grid. The environment is split into four sectors of $2\times2$ cells. We say that the agent is in sector $i$ if it is currently located in one of the cells within the sector. Symbol \verb|Si| indicates the agent is in sector \verb|i|. Symbols \verb|P| and \verb|D| indicate the passenger has been picked up or dropped off, respectively. The colors match abstract states to their corresponding sectors.}
        \label{fig:rm-split}
    \end{minipage}
    \caption{}
    \label{fig:taxi}
\end{figure*}

The EL contexts are represented by the location of the entities (passengers and their destinations) in the environment. This is a pair of row-column coordinates for each entity. The CM contexts are represented as a binary vector where each value indicates the existence of a wall at a certain position in the environment map. The PO contexts are represented by a single index for each passenger that is the position of the passenger in the pickup order. \cref{fig:contexts} shows two different contexts in the PD environment coupled with the EL context space.

\section{Experiment Reward Machines}\label{app:rm}
In the PO context space, there is an \rmGenFuncTerm of full resolution that generates simple \rmTerm{}s where the number of propositional symbols abstract states is equal to the number of passengers. The \rmGenFuncTerm template with variables \verb | i1,..., i2 | that defines the passenger pickup order is as follows:
\begin{verbatim}
Symbols:
    P1 - Passenger 1 has been picked up
    ...
    P5 - Passenger 5 has been picked up

Order-i1,i2,i3,i4,i5:
States - u0, ..., u5
Transitions - 
    (u0, not Pi1) --> next=u0;r=0
    (u0,     Pi1) --> next=u1;r=0
    ...
    (u4, not Pi5) --> next=u4;r=0
    (u2,     Pi5) --> next=u5;r=1
\end{verbatim}

For the EL context space, the entities may be anywhere on the map. This means that an \rmGenFuncTerm that differentiates between all contexts needs to use at least one symbol for every possible position. Because the agent can potentially visit any of these positions during an episode, we must have an abstract state for each position on the map and passenger status (waiting, picked up, delivered). Such an \rmTerm is about the same size as the entire MDP, and so it would be more beneficial to run VI on the MDP itself and guarantee a near-optimal policy rather than the \rmTerm. Similarly, the CM contexts determine the locations of obstacles, and so the \rmTerm must account for every possible transition in the map (which is also about the size of the MDP).

To overcome this issue, we use an \rmGenFuncTerm that groups adjacent cells into sectors and symbols to indicate the agent's presence in a specific sector. A transition between sectors is possible if a transition between cells of those sectors is possible. \cref{fig:rm-split} illustrates this in the PD environment. This significantly reduces the size of the generated \rmTerm{}s, but the \rmTerm can only account for transitions between sectors (partial resolution).

\section{Training Technical Details}\label{app:training}

We trained each \algDQN agent on a single CPU core for four million steps on \srcContexts and \tgtContexts. For the EL context space, we train on 100 source contexts and transfer to 200 target contexts. For the CM context space, we train 250 source contexts and transfer to 500 contexts. The GN environment contains only 36 possible entity positions and so with the EL context space, there are only 36 contexts. Thus, we train on 8 source contexts and transfer to 16 target contexts in this environment, leaving room for the sampled contexts to change between runs.

\algDQN hyperparameters were selected via grid search of potential candidates, followed by a manual search in areas of interest. To demonstrate \cPrep's resilience to hyperparameters, hyperparameters were chosen to optimize the \ctxCTL configuration and used globally across all other configurations. The Q-value estimator network is comprised of two hidden linear layers with Rectified Linear Unit (ReLU) activations, ending with a linear output layer of width equal to the number of agent actions. The parameters are optimized with Adam \citep{kingma_adam_2015}. \cref{tab:hypers} lists all hyperparameters. Random seeds were chosen to be the first 5 multiples of 42 for all experiments.

\begin{table}[t]
    \centering
        \caption{\algDQN hyperparameters for all baseline configurations.}
    \begin{tabular}{|l|l|}
        \hline
        \textbf{Parameter} & \textbf{Value} \\
        \hline
        Target Q-network update interval & 10,000 steps \\
        Exploration time & 50\% training duration \\
        replay buffer size & 1M \\
        Discount factor &  0.99 \\
        training frequency & every 4 steps \\
        gradient steps per training & 4 \\
        sample batch size & 32 \\
        Number of hidden layers & 2 \\
        Hidden layer width & 64 \\
        Hidden activations & ReLU \\
        learning rate & 0.0001 \\
        Adam $\beta_1$ & 0.9 \\
        Adam $\beta_2$ & 0.999 \\
        \hline
    \end{tabular}

    \label{tab:hypers}
\end{table}

\section{Evaluation}\label{app:eval}
Agent policies are evaluated as deterministic policies every 1\% of training completed. An additional evaluation occurs before training to account for zero-shot transfer. Each evaluation records the return acquired from running 50 episodes in randomly sampled contexts from the context set on which we are training. The source policy is also evaluated in \tgtContexts to allow analysis of generalization throughout training. The returns are averaged to estimate the expected discounted return, with which we calculate the transfer utilities.

Let $\history_{\policyFunc}$ be the training history of policy \policyFunc, that is, a mapping from the number of timesteps trained to the estimated expected discounted return of the policy at that time. For some training configuration (environment, context space, hyperparameters, etc.), denote the target and transferred policies by \tgtPolicy and \tsfPolicy, respectively. Denote some predetermined threshold by \thresh. We calculate our transfer utilities as follows:
\begin{equation*}
    \tttUtil(\policyFunc, \thresh) = \min\left\{\timeSym\middle|\history_{\policyFunc}(\timeSym)\geq\thresh\right\} 
    \tag{Time to Threshold}
\end{equation*}
\begin{equation*}
    \jsUtil(\tsfPolicy,\tgtPolicy) = \history_{\tsfPolicy}(0)
    \tag{Jumpstart}
\end{equation*}
\begin{equation*}
    \trUtil(\tsfPolicy,\tgtPolicy) = \frac{\auc(\history_{\tsfPolicy}) - \auc(\history_{\tgtPolicy})}{\auc(\history_{\tgtPolicy})}
    \tag{Transfer Ratio}
\end{equation*}
where \auc is the area under the curve, estimated by the average of all recorded values on the curve. To provide a single \ttt value that considers all thresholds, we also calculate utility $\tttAucUtil(\policyFunc) = \underset{\thresh}{\auc}(\tttUtil(\policyFunc, \thresh))$.

\section{\cPrep with \algDQN}\label{app:alg}
\cref{alg:train} shows the \algDQN algorithm with \cPrep integration. A textual description of the algorithm is available in \cref{sec:cprep}.

\begin{algorithm*}[t]
    \caption{Training \algDQN with \cPrep}\label{alg:train}

    \begin{algorithmic}
        \STATE {\bfseries Input:} $\policyFunc = \qFuncParameterized$ - initial \algDQN policy

        \STATE {\bfseries Input:} $\contextSpace$ - contexts set

        \STATE {\bfseries Input:} \itersSym - number of episodes to train

        \STATE {\bfseries Input:} $\rmGenFunc$ - \rmGenFuncTerm

        \STATE {\bfseries Input:} $\stateLabelFunc$ - \stateLabelFuncTerm

        \STATE $\replayBufferSym \gets$ empty experience replay buffer

        \FOR{$i \in [\itersSym]$}

            \STATE $\contextSym \gets sampleUniform(\contextSpace)$
            \STATE $\rmTup \gets$ generated \rmTerm for \contextSym
            \STATE $\valueFunc \gets$ value iteration on \rmTerm 
            \STATE $\rmAbsRewardFunc(\rmAbsStateSym,\rmPropSym) \gets \rmAbsRewardFunc(\rmAbsStateSym,\rmPropSym) + \discountFactor\valueFunc(\rmAbsTransFunc(\rmAbsStateSym,\rmPropSym)) - \valueFunc(\rmAbsStateSym)$ \COMMENT{reward shaping}
            \STATE $\rmAbsStateSym \gets \rmInitAbsStateSym$

            \FOR{each step in episode of $\mdpInContextSym$}

            \STATE $\rmPropSym^* \gets \underset{\rmPropSym \in 2^{\rmPropSet}}{\argmax} \rmAbsRewardFunc(\rmAbsStateSym,\rmPropSym) + \discountFactor \valueFunc(\rmAbsTransFunc(\rmAbsStateSym,\rmPropSym))$

            \STATE $\augStateSym \gets \tuple{\stateSym, \rmPropSym^*}$ \COMMENT{using augmented state space}
            
            \STATE $\qValuesSym \gets \qFuncParameterized(\augStateSym)$
            \STATE $\actionSym \gets \epsilon$-greedy$(\qValuesSym)$
            \STATE $\stateSym' \gets \mdpInContextSym$.step(\actionSym)

            \STATE $\rmAbsStateSym' \gets \rmAbsTransFunc(\rmAbsStateSym, \stateLabelFunc(\stateSym, \actionSym, \stateSym'))$
            \STATE $\rewardSym \gets \rmAbsRewardFunc(\rmAbsStateSym, \stateLabelFunc(\stateSym, \actionSym, \stateSym'))$ \COMMENT{using \rmTerm reward}

            \STATE ${\rmPropSym^*}' \gets \underset{\rmPropSym \in 2^{\rmPropSet}}{\argmax} \rmAbsRewardFunc(\rmAbsStateSym',\rmPropSym) + \discountFactor \valueFunc(\rmAbsTransFunc(\rmAbsStateSym',\rmPropSym))$
            \STATE $\augStateSym' \gets \tuple{\stateSym', {\rmPropSym^*}'}$

            \STATE $\replayBufferSym$.store$(\augStateSym, \actionSym, \rewardSym, \augStateSym')$

            \STATE $\augStateSym, \actionSym, \rewardSym, \augStateSym' \gets \replayBufferSym$.sampleBatch()
            \STATE $\qLossFunc \gets \underset{\text{batch}}{\sum}(\rewardSym +  \max_{\actionSym'}\left\{\qFuncParameterized(\augStateSym')[\actionSym']\right\} - \qFuncParameterized(\augStateSym)[\actionSym])^2$
            \STATE $\qParams \gets \nabla_{\qParams}\qLossFunc$
            \STATE $\stateSym, \rmAbsStateSym \gets \stateSym', \rmAbsStateSym'$
        \ENDFOR
        \ENDFOR
        \STATE {\bfseries Return}  $\qFuncParameterized$
    \end{algorithmic}

\end{algorithm*}

\section{Additional Experiments}\label{app:add-res}
\begin{figure*}[t]
    \centering
    \includegraphics[width=1\textwidth]{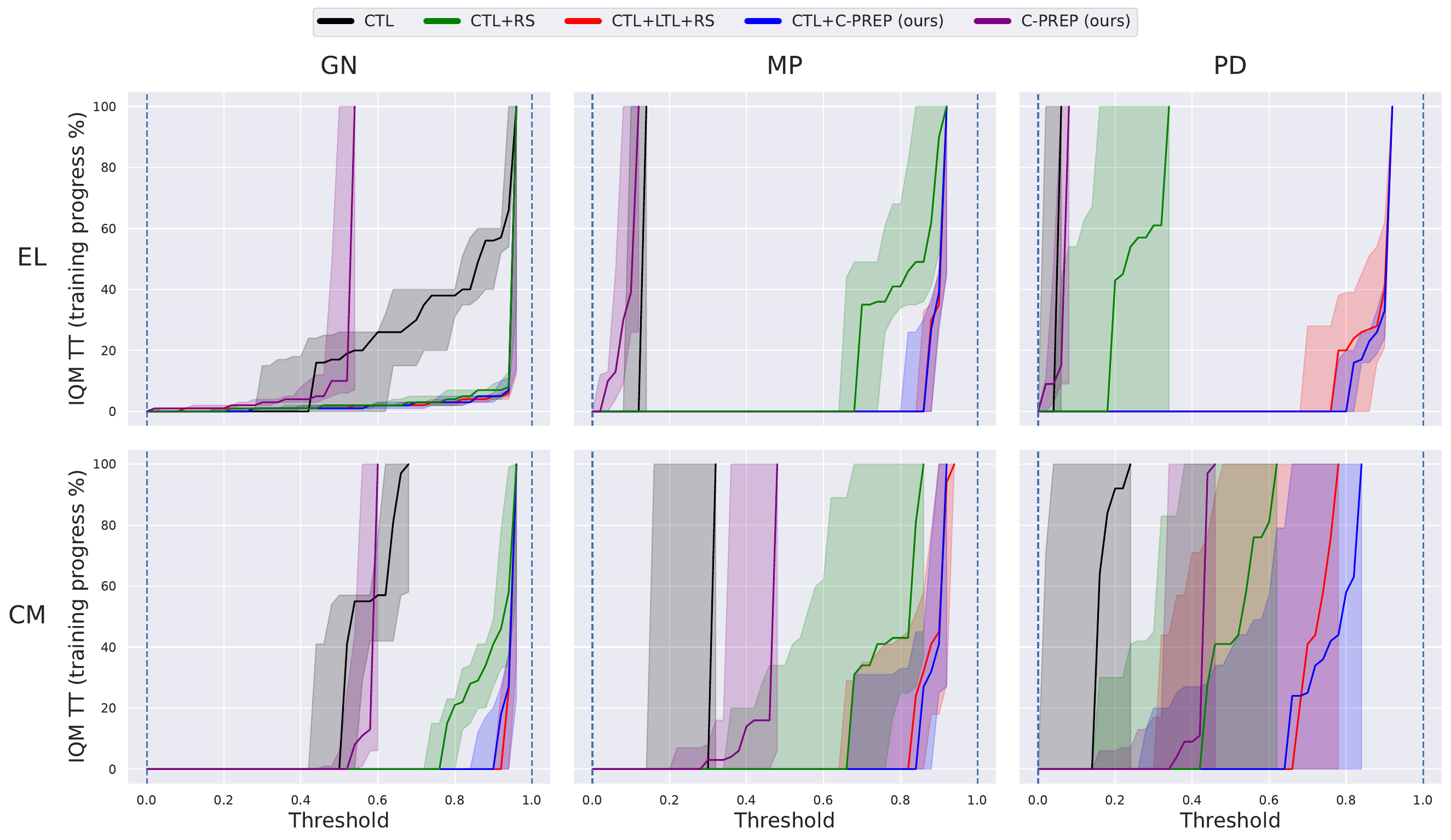}
    \caption{The IQM \ttt, measured in percentage of completed training, as a function of the threshold in environments GN, MP, PD, with doubley sized \srcContexts. Each color represents a different configuration, specified in the legend. The shaded areas indicate stratified bootstrap 95\% confidence intervals. Each row corresponds to a different context space, indicated to the left of the row. Each column corresponds to a different environment, indicated above the column.}
    \label{fig:res:doublesrc}
\end{figure*}

We perform two additional experiments. The first aims to show \cPrep's sample efficiency in terms of the number of source contexts on which it is trained. For this, we doubled the size of the source context set and rerun the experiments. IQM transfer utilities for this experiment are available in \cref{tab:doublesize}. \cref{fig:res:doublesrc} shows the \ttt performance over the achieved threshold for these settings in the GN, MP, and PD environments. As we can see, the performance gap between our method and \modLTL decreased when we doubled the size of \srcContexts. We conclude from this that \cPrep maintains its transfer performance even when the size of the source context set shrinks.

\begin{table*}[t]
    \centering
    \resizebox{1\textwidth}{!}{
    \bgroup
    \def\arraystretch{1.2}
    \begin{tabular}{|c|c|c?c|c|c|c|c|}
\hline
 Utility   & Context Space   & Environment   & CTL                            & CTL+RS             & CTL+LTL+RS                     & CTL+C-PREP (ours)    & C-PREP (ours)                  \\
\Xhline{4\arrayrulewidth}
\parbox[t]{2mm}{\multirow{7}{*}{\rotatebox[origin=c]{90}{$\ttt_{\auc}$}}}   & \multirow{3}{*}{EL} & GN   & 19.07 $\pm$ 3.10                  & 6.26 $\pm$ 0.51       & {\bf 5.92 $\pm$ 0.27}             & 6.13 $\pm$ 0.70         & 43.95 $\pm$ 4.74                  \\
    &   & MP   & 86.27 $\pm$ 7.70                  & 18.63 $\pm$ 6.06      & {\bf 10.03 $\pm$ 0.69}            & 10.24 $\pm$ 0.92        & 87.03 $\pm$ 1.74                  \\
    &   & PD   & 94.12 $\pm$ 15.42                 & 68.17 $\pm$ 24.76     & 12.96 $\pm$ 3.58                  & {\bf 11.70 $\pm$ 0.58}  & 87.92 $\pm$ 1.39                  \\
    & \multirow{3}{*}{CM}    & GN   & 37.99 $\pm$ 1.16                  & 10.22 $\pm$ 2.15      & {\bf 5.40 $\pm$ 0.88}             & 5.52 $\pm$ 1.09         & 34.74 $\pm$ 4.14                  \\
    &     & MP   & 67.83 $\pm$ 13.45                 & 20.97 $\pm$ 16.25     & 10.96 $\pm$ 6.24                  & {\bf 10.85 $\pm$ 4.82}  & 49.75 $\pm$ 9.38                  \\
    &     & PD   & 78.22 $\pm$ 22.00                 & 44.16 $\pm$ 23.82     & 27.30 $\pm$ 27.32                 & {\bf 22.16 $\pm$ 19.23} & 48.33 $\pm$ 15.64                 \\
    & PO    & ON   & 98.04 $\pm$ 0.00                  & 97.10 $\pm$ 16.40     & 96.31 $\pm$ 6.65                  & 23.29 $\pm$ 12.84       & {\bf 22.69 $\pm$ 1.48}            \\
 \Xhline{4\arrayrulewidth}
\parbox[t]{2mm}{\multirow{7}{*}{\rotatebox[origin=c]{90}{\js}}}   & \multirow{3}{*}{EL} & GN   & {\bf 0.42 $\pm$ 0.15}             & 0.21 $\pm$ 0.06       & 0.08 $\pm$ 0.15                   & 0.26 $\pm$ 0.12         & 0.00 $\pm$ 0.02                   \\
         &   & MP   & 0.13 $\pm$ 0.08                   & 0.69 $\pm$ 0.04       & {\bf 0.87 $\pm$ 0.02}             & 0.86 $\pm$ 0.04         & 0.03 $\pm$ 0.02                   \\
         &   & PD   & 0.06 $\pm$ 0.16                   & 0.19 $\pm$ 0.23       & 0.78 $\pm$ 0.09                   & {\bf 0.80 $\pm$ 0.03}   & 0.00 $\pm$ 0.02                   \\
         & \multirow{3}{*}{CM}    & GN   & 0.52 $\pm$ 0.06                   & 0.76 $\pm$ 0.05       & {\bf 0.93 $\pm$ 0.01}             & 0.91 $\pm$ 0.04         & 0.53 $\pm$ 0.06                   \\
         &     & MP   & 0.31 $\pm$ 0.13                   & 0.66 $\pm$ 0.19       & 0.83 $\pm$ 0.11                   & {\bf 0.84 $\pm$ 0.09}   & 0.29 $\pm$ 0.14                   \\
         &     & PD   & 0.12 $\pm$ 0.23                   & 0.43 $\pm$ 0.23       & {\bf 0.67 $\pm$ 0.29}             & 0.66 $\pm$ 0.28         & 0.36 $\pm$ 0.17                   \\
         & PO    & ON   & 0.00 $\pm$ 0.00                   & 0.00 $\pm$ 0.00       & 0.00 $\pm$ 0.00                   & 0.74 $\pm$ 0.31         & {\bf 0.78 $\pm$ 0.01}             \\
 \Xhline{4\arrayrulewidth}
\parbox[t]{2mm}{\multirow{7}{*}{\rotatebox[origin=c]{90}{\tr}}}   & \multirow{3}{*}{EL} & GN   & \textit{-0.02 $\pm$ 0.07} & {\bf 0.14 $\pm$ 0.04} & 0.08 $\pm$ 0.03                   & 0.08 $\pm$ 0.03         & 0.08 $\pm$ 0.04                   \\
         &   & MP   & \textit{-0.98 $\pm$ 0.12} & {\bf 0.56 $\pm$ 0.36} & 0.33 $\pm$ 0.07                   & 0.27 $\pm$ 0.07         & 0.11 $\pm$ 0.15                   \\
         &   & PD   & \textit{-0.98 $\pm$ 0.03} & {\bf 0.61 $\pm$ 0.64} & 0.43 $\pm$ 0.48                   & 0.32 $\pm$ 0.05         & \textit{-0.00 $\pm$ 0.14} \\
         & \multirow{3}{*}{EL}    & GN   & \textit{-0.05 $\pm$ 0.07} & {\bf 0.23 $\pm$ 0.02} & 0.13 $\pm$ 0.01                   & 0.14 $\pm$ 0.01         & 0.09 $\pm$ 0.04                   \\
         &    & MP   & \textit{-0.75 $\pm$ 0.30} & {\bf 0.36 $\pm$ 0.07} & 0.30 $\pm$ 0.09                   & 0.28 $\pm$ 0.05         & 0.10 $\pm$ 0.04                   \\
         &     & PD   & \textit{-0.29 $\pm$ 0.37} & {\bf 0.36 $\pm$ 0.31} & 0.11 $\pm$ 0.24                   & 0.26 $\pm$ 0.06         & 0.08 $\pm$ 0.04                   \\
         & PO    & ON   & 0.00 $\pm$ 0.00                   & 0.45 $\pm$ 0.88       & \textit{-0.80 $\pm$ 7.12} & {\bf 0.83 $\pm$ 0.34}   & 0.33 $\pm$ 0.08                   \\
\hline
\end{tabular}
\egroup
}
\caption{IQM and standard deviation transfer utilities  for all tested configurations (environment-context space pairs) using \ctxCTL as the base context representation trained on doubly sized source context set before transfer, aggregated over all seeds. The best results for each CMDP (row) are marked in bold. Negative \tr values that indicate non-beneficial transfer are italicized.}
    \label{tab:doublesize}
\end{table*}

\begin{figure*}[t]
    \centering
    
    \includegraphics[width=0.7\textwidth]{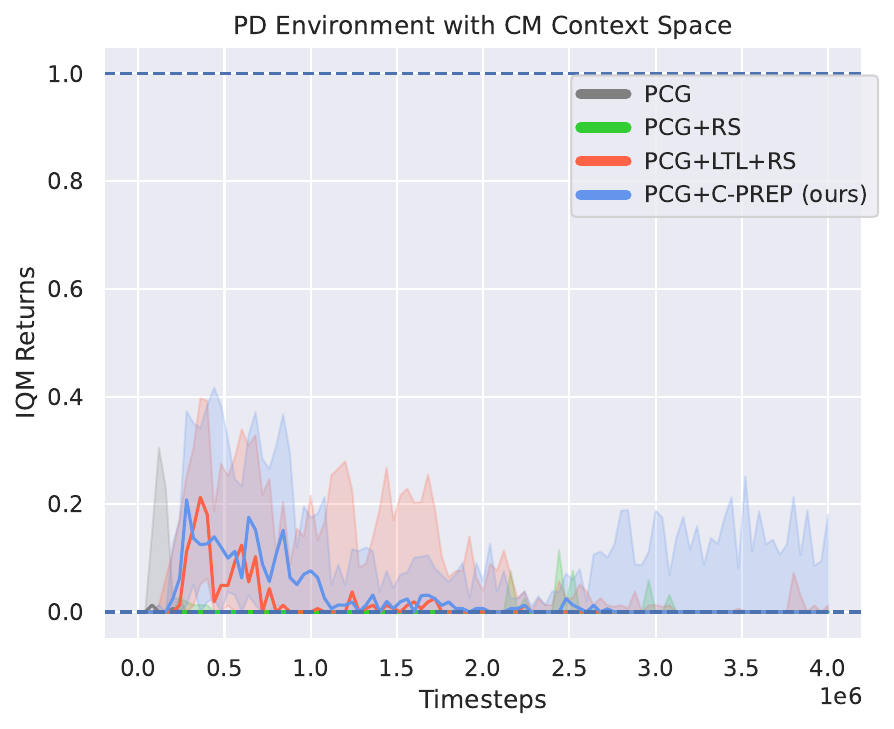}
    
    \caption{The IQM performance of the source policy, evaluated on the target context set. in environment PD and context space CM with \ctxPCG configurations. Each color represents a different configuration, specified in the legend. The shaded areas indicate stratified bootstrap 95\% confidence intervals. Each row corresponds to a different context space, indicated to the left of the row. Each column corresponds to a different environment, indicated above the column.}
    \label{fig:res:ohe_test}
\end{figure*}

The second experiment tested the generalization capabilities of \cPrep using \ctxPCG representations. As we saw in \cref{sec:eval:res} and analyzed in \cref{sec:eval:disc}, we witness severe overfitting when using \ctxPCG. However, we find that using \rmTerm information is not completely futile. Figure \cref{fig:res:ohe_test} shows the performance of the source policy during training in \srcContexts, evaluated in \tgtContexts in the PD environment with the CM context space. We observe that in the first 30\% of training, there is a spike in performance for \modLTL{} +\modRS or \cPrep. These are the only configurations that use additional context representations from \rmTerm information to augment the contextual \ctxPCG input. The performance spike is far too high to be coincidental since otherwise, other configurations should also display this phenomenon. This begs the question ``What is learned before overfitting occurs and how can we preserve this knowledge''? Furthermore, we still do not have a clear explanation as to why specifically \rmTerm information causes this spike. We believe this has implications for representation learning, where agents learn latent representations of the state space in a disentangled manner from the policy.

\section{Multi-Taxi Environment}\label{app:multi-taxi}
\begin{figure*}[t]
    \centering
        \includegraphics[width=0.5\textwidth]{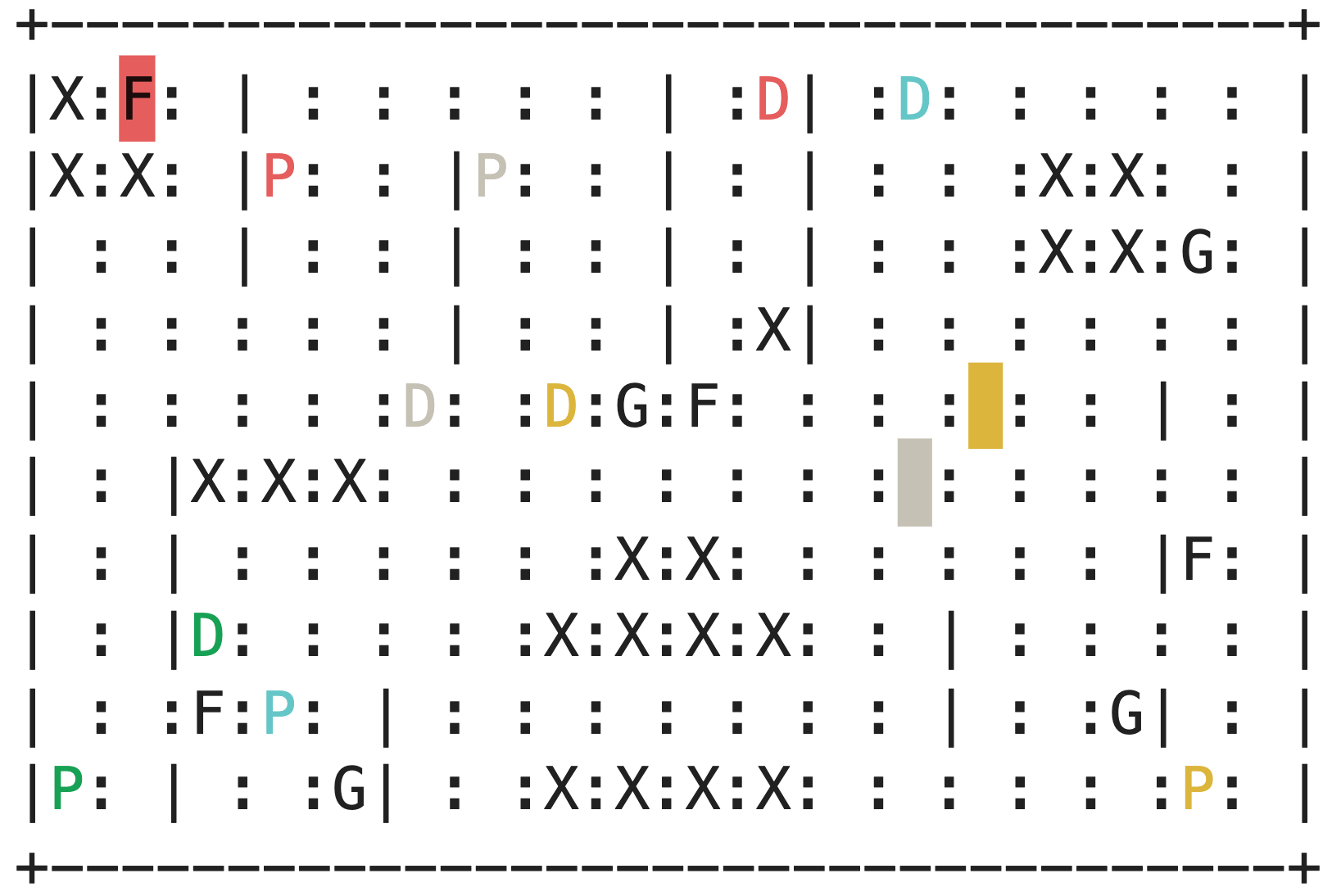}
    \caption{A visualization of the Multi-Taxi environment. Colored rectangles are taxi agents, and P and D symbols indicate the location of a passenger and its corresponding destination (respectively). 'X' and '|' values indicate different kinds of obstacles. 'F' and 'G' are two different kinds of fuel stations.}
    \label{fig:screenshot}
\end{figure*}

{\bf Multi-Taxi} is a highly configurable multi-agent environment, based on the OpenAI gym taxi environment \cite{brockman_openai_2016}, which adheres to the PettingZoo API \citep{terry_pettingzoo_2021}. \cref{fig:screenshot} shows a visualization of the environment. The environment's configurable features allow the user to set the number of passengers and taxis, the taxi's capacity and fuel requirements, the actions' stochasticity, the sensor function, and more. We note that while multi-taxi is natively a multi-agent environment, we explore it in the single-agent setting. By leveraging the domain's customizability we define seven different environment settings of varying complexity levels based on three context spaces changing different aspects of the environment between tasks. The code can be found at \url{https://github.com/CLAIR-LAB-TECHNION/multi_taxi}.

\section{Acronyms}\label{app:acronyms}
\cref{tab:glossary} is a glossary for frequently used terms in this work. The table is divided into three sections using thick horizontal lines. The top section contains general terms. The middle section contains names of context representations used in our experiments. The bottom section contains the environments and context spaces used in our experiments.

\begin{table*}[t]
    \centering
    \begin{tabular}{|p{0.0675\textwidth}|p{0.3\textwidth}|p{0.6325\textwidth}|}
        \hline
        \textbf{Abbv} & \textbf{Full Term} & \textbf{Description} \\
        
        \Xhline{4\arrayrulewidth}
        
        RL & reinforcement learning & A method for agent learning through experiencing the world, acting within it, and receiving rewards for achieving certain states or state transitions. \\
        
        \hline
        DRL & deep reinforcement learning & RL using deep learning methods. \\

        \hline
        TL & transfer learning & The improvement of learning a new task through the transfer of knowledge from a related task that has already been learned. \\

        \hline
        \MDP & Markov decision process & A world model used commonly in RL, defined as a 5-tuple \mdpTup. \\

        \hline
        \CMDP & contextual Markov decision process & A collection of \MDP{}s identified by contexts, defined as a 4-tuple \cmdpTup. \\

        \hline
        VI & value iteration & A dynamic programming algorithm for estimating the optimal state-value function in an \MDP. \\

        \hline
        \rmTerm & reward machine & A state machine abstractions of an MDP, defined as a 3-tuple \rmTup. \\

        \Xhline{4\arrayrulewidth}
        
        \ctxCTL & controllable context representation & A context representation that includes the necessary information to generate the MDP. \\

        \hline
        \ctxPCG & procedural content generation context representation & A context representation that conceals the MDP variables and only reveals information about the context identity. \\

        \hline
        \modLTL & last transition label & A configuration of the learning algorithm using the \rmTerm transition label mapped from the previous \MDP transition. \\

        \hline
        \modDTL & desired transition label & A configuration of the learning algorithm using the transition label that leads to an optimal transition within the \rmTerm. \\

        \hline
        \modRS & reward shaping & A configuration of the learning algorithm using potential-based reward shaping based on an \rmTerm. \\

        \hline
        \cPrep & contextual pre-planning & The method presented in this work, whose configuration is \modDTL{}+\modRS.  \\

        \Xhline{4\arrayrulewidth}
        
        GN & Grid Navigation & An environment where the agent must navigate to a specific location and declare ``done''. \\

        \hline
        MP & multi points-of-interest & An environment where the agent must navigate to multiple locations in no particular order, declaring arrival at each location. \\

        \hline
        PD & pick-up and drop-off & An environment where the agent must pick up and drop off passengers at their desired destinations. \\

        \hline
        ON & ordered navigation & An environment where the agent must navigate to multiple locations in a specific order, declaring arrival at each location. \\

        \hline
        EL & entity location & A context space where contexts affect the location of entities on the map (eg., points-of-interest and passenger destinations)\\

        \hline
        CM & changing map & A context space where contexts affect the location of obstacles on the map. \\

        \hline
        PO & points-of-interest order & A context space for the ON environment where contexts affect the order by which the points-of-interest must be visited. \\
        
        \hline
    \end{tabular}
\caption{A glossary of acronyms used frequently in this work.}
    \label{tab:glossary}
\end{table*}

\end{document}